\documentclass{article}
\usepackage{a4wide}
\usepackage{graphicx} 

\usepackage[T1]{fontenc}
\usepackage[utf8]{inputenc}
\usepackage[english]{babel}
\usepackage{lmodern}
\usepackage{algorithm}
\usepackage{algpseudocode}
\usepackage{amsmath,amssymb,amsthm}
\usepackage{tikz}
\usepackage{graphicx}
\usepackage{caption}
\usepackage{subcaption}
\usepackage{xcolor}
\usepackage{cite}
\usepackage[final]{hyperref} 
\usepackage[autostyle,italian=guillemets]{csquotes}
\usepackage{fancyhdr}

\newcommand*{\m}{\textrm{max}}

\title{A Multiscale Kinetic Framework for Image Segmentation: From Particle Systems to Continuum Models}
\author{
		Horacio Tettamanti\thanks{\texttt{horacio.tettamanti01@universitadipavia.it}}\\
		{\small	Department of Mathematics ``F. Casorati''} \\
		{\small University of Pavia, Italy} \\
		\\
        Giulia Guicciardi\thanks{\texttt{giulia.guicciardi01@universitadipavia.it}}\\
		{\small	Department of Mathematics ``F. Casorati''} \\
		{\small University of Pavia, Italy} \\
		\\
		 Mattia Zanella\thanks{\texttt{mattia.zanella@unipv.it}} \\
		{\small	Department of Mathematics ``F. Casorati''} \\
		{\small University of Pavia, Italy}
		}

\begin{document}
\maketitle

\begin{abstract}

In this work, we present a multiscale kinetic framework for consensus-based image segmentation. By interpreting an image as a system of interacting particles, each pixel is characterised by its spatial position and an internal feature encoding color information. We introduce a coupled interaction scheme governing the evolution of particles in both position and feature spaces, from which we derive a kinetic formulation for the particle density in the space-feature domain combining transport, aggregation, and diffusion effects. Furthermore, through a suitable scaling, we obtain a first-order macroscopic model describing the evolution of the fraction of pixels carrying information on the fraction of pixels having a certain feature. Based on this reduced-complexity model, we present a data-oriented approach where we make use of particle-based optimisation techniques for the accurate segmentation of images. Numerical tests show the effectiveness of the proposed framework and its robustness under different noise conditions.

\end{abstract}

\section{Introduction}

Image segmentation is a fundamental task in computer vision and image processing, consisting in partitioning an image into meaningful regions such that pixels within each region share similar characteristics, including color variations, intensity values, texture patterns, and spatial proximity. The objective is to simplify the image representation, thereby facilitating its analysis. This process plays a crucial role in a wide range of applications, particularly in medical imaging and clinical research, where accurate identification and delineation of anatomical structures, lesions, and other relevant features are essential for diagnosis, treatment planning, and disease monitoring \cite{ABJT,ABCDMPSSSS,CDA}.

Over the past decades, a wide range of computational techniques has been developed to address the problem of image segmentation. Among them, deep learning-based approaches have emerged as the most widely established methods \cite{IS,KLPT,RFB,ZRTL}. These techniques have demonstrated remarkable accuracy in delineating regions of interest (ROI), particularly when large annotated datasets are available. However, their performance strongly depends on the quality and quantity of training data, which constitutes a significant limitation, especially in medical imaging, where annotated datasets are often scarce. As a way to overcome this limitation, transfer learning techniques have been proposed, which leverage pre-trained models on large datasets to improve performance on smaller, domain-specific datasets \cite{KKJP}. 

In addition to supervised learning approaches, clustering techniques have also been explored for image segmentation \cite{CA,HS,JMF,K}. These methods partition an image into distinct regions according to similarity criteria. Their main advantage lies in the fact that they do not require annotated data, making them particularly suitable in scenarios where labeled datasets are scarce.

Building on this perspective, recent work has focused on the development of a kinetic approach to image segmentation based on consensus-based models for opinion dynamics~\cite{CTZ,CPLFZ}. In this framework, an image is interpreted as a system of pixels unlike classical interacting particles, where each pixel is characterized by its spatial position and an internal feature encoding color information. A virtual interaction scheme, inspired by consensus-type dynamics, is introduced, whereby pixels cluster according to spatial and feature proximity. In this initial formulation, the feature variable remains static in time. As a result, the asymptotic distribution of the system is characterized by a finite number of clusters, determined by the model parameters. Notably, unlike most unsupervised techniques, this approach does not require prescribing the number of clusters a priori.

Within this setting, the kinetic theory of multi-agent systems provides a rigorous connection between microscopic interaction rules and macroscopic behaviour~\cite{AP,AHP,APZ,APZ2,CS,PT,T,Z}. At the mesoscopic level, the density function in the space-feature domain satisfies a Boltzmann-type equation, which, under an appropriate quasi-invariant scaling, converges to a Fokker-Planck-type equation. This formulation enables the use of well-established numerical methods, such as Direct Simulation Monte Carlo and asymptotic-preserving schemes, to accurately capture the long-time behaviour of the system~\cite{B,BI,DP,PR,PR2,PZ}.

The optimal set of model parameters are determined through an optimization procedure in which a loss function measures the discrepancy between the predicted segmentation mask and the ground truth. The impact of the choice of loss function has been studied in~\cite{CTZ}. This optimisation problem is challenging due to the non-convexity of the loss function and the high computational cost associated with evaluating segmentation masks for different parameter configurations.

In this paper, we present an extension of the previous model inspired by the work in \cite{PTTZ}. Based on the extensive literature on the macroscopic description of kinetic systems, we introduce a multiscale kinetic framework that provides a macroscopic view of the evolution of the relevant quantities, enabling the construction of a reduced-complexity model for image segmentation \cite{BEG,CFRT,HT,OCV}. This, in turn, allows for a significant reduction in computational cost, thereby enabling the exploration of different optimization strategies for parameter selection. In particular, we adopt a data-driven approach based on Consensus-Based Optimization (CBO), which is capable of accurately identifying regions of interest under varying noise conditions \cite{CBE,CJLZ,BBGRTV}.

The remainder of the manuscript is organized as follows. In Section \ref{sect:kinetic}, we introduce the multiscale kinetic framework for image segmentation, including the microscopic scheme and its corresponding kinetic description. Section \ref{sect:macro} is devoted to the derivation of a first-order macroscopic model for the evolution of the relevant quantities. In Section \ref{sect:num} we present the consistency of the proposed method through a number of numerical experiments and the application of the presented framework to the problem of image segmentation. We finally conclude with some remarks and future research directions in Section \ref{sect:conclusions}.

\section{Multiscale Kinetic Framework for Image Segmentation}\label{sect:kinetic}

Inspired by recent developments in consensus-based models for image segmentation, we present a novel approach based on a multiscale kinetic framework. Our model builds on the framework proposed in \cite{PTTZ}, where a kinetic model for opinion dynamics is introduced in which the opinion variable is coupled with an additional feature representing individuals’ preferences with respect to a given topic. The authors derive a hydrodynamic description for the evolution of the conserved quantities of the interaction process. The resulting macroscopic formulation provides a reduced-complexity model that captures large-scale emerging dynamics, such as consensus and polarisation, through reduced complexity models describing the evolution of the moments of the kinetic particle density. Building on this framework, we formulate an image segmentation approach by interpreting the opinion variable as the spatial variable and the preference variable as the grey-level intensity of the pixels. 

\subsection{Microscopic Interaction Dynamics}
Each pixel of the image is characterised by its position $\mathbf{x}_i \in \Omega \subset \mathbb R^2$ and by an internal feature $c_i \in [0,1]^d$, $d\ge1$. 
At the microscopic level, we consider the following virtual scheme for pixels interaction
\begin{equation}
\begin{cases}
\displaystyle \frac{dc_i}{dt} &= -(c_i - \alpha_{\Delta_1}(\mathbf{x}_i))\,\phi(c_i), \\
d\mathbf{x}_i &= \dfrac{1}{N} \displaystyle \sum_{j=1}^N  P_{\Delta_2}(c_i,c_j)\,(\mathbf{x}_j - \mathbf{x}_i)\,dt
+ \sqrt{2\sigma^{2}(\mathbf{x}_i,c_i)}\, d\mathbf{W}_i,  
\end{cases}
\label{eq:micro_scheme}
\end{equation}
where $\alpha_{\Delta_1}(\cdot): \mathbb R^2 \to [0,1]^d$ is defined as
\begin{equation}
\label{eq:alpha}
\alpha_{\Delta_1}(\mathbf{x}_i)= \frac{1}{S_i}\sum_{j\in S_i}c_j
\end{equation}
where $S_i = \sum_{j=1}^N \chi( |\mathbf{x}_j - \mathbf{x}_i|<\Delta_1)$ defines a ball around the $i$th pixel of radius $\Delta_1>0$. We observe that $\alpha_{\Delta_1}(\cdot)$ encodes the local average feature value, while $\phi(c_i): [0,1]^d \to [0,1]$ acts as a modulation function controlling the strength of aggregation toward $\alpha_{\Delta_1}$, so that the interaction vanishes at the boundary and boundary values remain unaffected by the dynamics. In particular, a possible form of this function is 
\[
\phi(c_i) = \frac{d}{2} - \sum_{\ell = 1}^d \left|c_i^{(\ell)}-\frac{1}{2}\right|,
\]
being $c^{(\ell)}_i$ the $\ell$th component of $c_i \in [0,1]^d$.

The second equation in \eqref{eq:micro_scheme} describes the evolution of the position of the pixels. We introduced $P_{\Delta_2}(\cdot,\cdot): [0,1]^d \times [0,1]^d \to [0,1]$ modulating the interaction between pairs of pixels and depending on their feature values. Consistently with \cite{CPLFZ,CTZ} we will consider
\begin{equation}
\label{eq:P}
P_{\Delta_2}(c_i,c_j) = \chi\!\bigl(|c_j - c_i| < \Delta_2\bigr),\end{equation} such that two pixels interact if their features are sufficiently homogeneous. The function $P_{\Delta_2}(\cdot,\cdot)$ is of bounded-confidence-type \cite{HK} and depends on the confidence threshold $\Delta_2 \in[0,1]$. In \eqref{eq:micro_scheme} we introduced the set of $N$ independent Wiener processes $\{\textbf{W}_i\}_{i=1}^N$ weighted by $\sigma^2$ which may depend on both $c_i \in [0,1]$ and $\mathbf{x}_i\in \mathbb R^2$. 


In this setting, the dynamics in space are governed by an aggregation mechanism acting between particles that are sufficiently close in the feature space, together with a stochastic contribution whose intensity is controlled by the parameter $\sigma$. 
Furthemore, the evolution of the feature variable is dictated by a transport term that drives the feature value of each particle towards a local average $\alpha(\mathbf{x}_i)$, which is computed as the mean feature value of the particles that are sufficiently close in space. The speed of this transport is modulated by a function $\phi(c_i) = \frac{1}{2} - |c_i - \frac{1}{2}|$. The advantage of this choice of the virtual microscopic scheme is that the evolution in the space and feature are decoupled. As we will present in the following sections, the advantage of this is that it will allow us to obtain a macroscopic description for the evolution of the relevant macroscopic quantities.

\subsection{Binarisation Mechanism}

The asymptotic distribution of the interaction dynamics described in the previous section consists of a finite number of homogeneous regions, where pixels share similar feature values. To obtain segmentation masks, it is necessary to further drive these feature values toward a prescribed set of target states. In this work, we focus on the binary case, where the feature values are steered toward either $0$ or $1$. This setting is particularly relevant in medical image segmentation, where the objective is to identify regions of interest—such as lesions or anatomical structures—represented by binary masks. To this end, we introduce a binarization mechanism in the feature space, which can be described by the following dynamics.

\begin{equation}
\begin{aligned}
\frac{d}{dt}c_i &= - \nabla_{c}V(c_i) 
\end{aligned}
\label{eq:binarization}
\end{equation}
being $V(\cdot)$ a potential whose form is
\[
V(c) =\, \sum_{i=1}^d A_{c_{\m},i}\, c_i^{\alpha_i}(1-c_i)^{\beta_i}, \qquad c \in [0,1]^d,\alpha_i,\beta_i>0
\]
where $d$ represents the dimension of the feature space. 
The feature potential depends on the value $c_\m\in [0,1]^d$  representing the maximum value of a double-well potential. 
The pixels characterized by a feature level $c_i> c_{\m,i}$ are driven toward the vector $1$ in the $i$th component, while the pixels characterized by a feature level $c_i<c_{\m,i}$ are driven toward $0$ in the $i$th component.  We introduce a normalization constant $A_{\tilde{c}}$ to ensure that the maximum value of the potential, attained at $c=\tilde{c}$, is the same for all admissible choices of $\tilde{c}$. In particular, we impose
\[
V(c_{\m}) = \frac{d}{4},
\]
which yields
\[
A_{c_{\m},i} = \frac{1}{4\,c_{\m,i}^{\,\alpha_i}(1-c_{\m,i})^{\beta_i}}, \qquad i = 1,\dots,d.
\]

More generally, the parameters $\alpha_i$ and $\beta_i$ are chosen so as to satisfy the following structural conditions on the potential. Upon imposing that the potential attains a maximum at $c = c_\m$ we get
\begin{equation}\label{eq:rel}
    \frac{\alpha_i}{\beta_i} = \frac{c_{\m,i}}{1 - {c_{\m,i}}}, \qquad i = 1,\dots,d. 
\end{equation}
Moreover, we impose that
\begin{equation}
\label{eq:cond2}
\nabla_c V(c)\big|_{\partial[0,1]^d} = 0,
\end{equation}
so that pixels such that $c_i \in\{0,1\}$ remain stationary and thus preventing the particles from escaping the domain. The gradient of the potential reads
\begin{equation}
    \partial_{c_i} V = c_i^{\alpha_i-1}(1-c_i)^{\beta_i-1} \left( \alpha_i (1-c_i) - \beta_i c_i\right),
\end{equation}
from which we get that \eqref{eq:cond2} is met provided 
\[
\alpha_i > 1, \qquad \beta_i > 1.
\]
Thanks to the relation between $\alpha_i$ and $\beta_i$, $i=1,\dots,d$ defined in equation~\ref{eq:rel}, this translates into imposing
$$\alpha \frac{1 - {c_{\m,i}}}{{c_{\m,i}}} > 1,$$
and $\alpha_i>1$. Hence, for a given value of ${c}_{\m,i}$, the parameters $\alpha_i$ must satisfy
\begin{equation}
    \begin{cases}
        \alpha_i > 1, & \text{if } {c}_{\m,i} < \frac{1}{2}, \\[6pt]
        \alpha_i > \dfrac{{c}_{\m,i}}{1 - {c}_{\m,i}}, & \text{if } {c}_{\m,i} > \frac{1}{2}.
    \end{cases}
\end{equation}

\subsection{Kinetic Formulation of the Model}

In order to describe our dynamics from a kinetic point of view we set up the binary interaction between a pair of generic particles $(\mathbf{x},c)\in \Omega \times [0,1]^d$ and $(\mathbf{x}_*,c_*)\in \Omega \times [0,1]^d$. To this end, we may proceed as in \cite{AP,CPLFZ} to obtain the following binary scheme characterising the space dynamics
\begin{equation}
\label{eq:binary-interaction}
\begin{split}
\mathbf{x}'  &= \mathbf{x}
+ \varepsilon P_{\Delta_2}(c,c_*)(\mathbf{x}_*-\mathbf{x})
+ \sqrt{2\sigma^2(\mathbf{x},c)}\,\eta,\\[4pt]
\mathbf{x}_*' &= \mathbf{x}_*
+ \varepsilon P_{\Delta_2}(c_*,c)(\mathbf{x}-\mathbf{x}_*)
+ \sqrt{2\sigma^2(\mathbf{x}_*,c_*)}\,\eta_*.
\end{split}
\end{equation}
where $\eta,\eta_*$ are two independent 2D random variables  such that
\[
\left\langle\eta\right\rangle=\left\langle\eta_*\right\rangle=0,\qquad 
\left\langle\eta^2\right\rangle=\left\langle\eta_*^2\right\rangle=1
\]
In \eqref{eq:binary-interaction} we introduced the local relevance of the diffusion $\sigma^2(\mathbf{x},c)\ge0$ that can be chosen in such a way $\mathbf{x}',\mathbf{x}_*' \in \Omega$. Possible structures of such a function are $\sigma^2(\mathbf{x},c) = D(c) d^p(\mathbf{x})$, $p\ge1$, and $D(c) = 0$ if $c \in \partial[0,1]^d$, being $d(\mathbf{x}) = \textrm{dist}(\mathbf{x},\partial\Omega)$, see e.g. \cite{CPLFZ,CTZ,T}. 

Similarly, we write the update in the feature space. We notice that in this case the dynamics is linear in both cases so we do not consider the interaction between a pair of particles but rather the linear transport and binarization steps separately
\begin{equation}
\label{eq:binary-binarization}
\begin{split}
c'  &= c   - \varepsilon(c-\alpha_{\Delta_1}(\mathbf{x}))\,\phi(c),\\[4pt]
c''  &= c   - \varepsilon \nabla_c V(c)
\end{split}
\end{equation}

We can observe that the first equation in \eqref{eq:binary-binarization} can be rewritten as follows
\[
c^\prime = (1-\varepsilon \phi(c))c + \varepsilon \alpha_{\Delta_1}(\mathbf{x})\phi(c).
\]
Therefore $c'\in [0,1]^d$ provided $0\le \varepsilon\phi(c)\le 1$ and $\alpha_{\Delta_1}(\mathbf{x}) \in [0,1]^d$. We remark that the maximum value of $\phi(c)$ is $d/2$ therefore the value of $\varepsilon$ should be consistent such that the previous relation is satisfied. Furthermore, assuming $\alpha_i,\beta_i>1$ we get $|\partial_{c_i} V|\le M_i<M<+\infty$, $\nabla_c V$ Lipschitz continuous on $[0,1]^d$ and vanishing at the boundaries of the domain. Hence, there exist $\varepsilon>0$ such that $c''\in [0,1]^d$ provided $c \in [0,1]^d$. 

We now introduce the distribution function $f(\mathbf{x},c,t) = f(\mathbf{x},c,t) : \mathbb{R}^2 \times [0,1]^d \times \mathbb{R}_+ \rightarrow \mathbb{R}$ where $f(\mathbf{x},c,t)d\mathbf{x}\,dc$ represents the fraction of particles with position in the physical space between $[\mathbf{x},\mathbf{x}+d\mathbf{x})$ and with feature value between $[c,c+dc)$ at time $t>0$.
The time evolution of the density function $f(\mathbf{x},c,t)$ under the binary interaction scheme detailed in \eqref{eq:binary-interaction}-\eqref{eq:binary-binarization} can be described by the following Boltzmann-type equation 

\begin{equation}
\partial_t f(\mathbf{x},c,t) = \theta_S Q_{S}(f,f)(\mathbf{x},c,t) + \theta_F Q_{F}(f)(\mathbf{x},c,t) + \theta_B Q_{B}(f)(\mathbf{x},c,t) 
\label{eq:Boltzmann}
\end{equation}
where $\theta_S,\theta_F,\theta_B>0$ and we introduced the kinetic operators 
\begin{equation}
\label{eq:operators}
\begin{split}
&Q_{S}(f,f)(\mathbf{x},c,t) = \\
&\quad\left\langle 
\int_{\Omega}\int_{[0,1]^d}\frac{1}{'J_S}\left(
f(\mathbf{x}',c,t)\,f(\mathbf{x}_*',c_*,t)
-f(\mathbf{x},c,t)\,f(\mathbf{x}_*,c_*,t)
\right)
\, dc_*\, d\mathbf{x}_*
\right\rangle, \\
&Q_{F}(f)(\mathbf{x},c,t) = 
\frac{1}{'J_F}
f(\mathbf{x},c',t)
-
f(\mathbf{x},c,t),\\
&Q_{B}(f)(\mathbf{x},c,t) =
\frac{1}{''J_{B}}\, f(\mathbf{x},c'',t)
-
f(\mathbf{x},c,t)
\end{split}
\end{equation}
where $Q_{S}(\cdot,\cdot)$ accounts for the interaction in space, while $Q_{F}(\cdot,\cdot)$ is the operator encapsulating for the interaction in the feature space given by the interactions scheme defined  in~\ref{eq:binary-interaction}. Moreover, $Q_{B}(\cdot)$ represents the operator associated with the binarization interaction shown in \ref{eq:binary-binarization}. The introduced operators are weighted by the coefficients $\theta_S,\theta_F,\theta_B>0$ tuning the frequency of the updates. 

In \eqref{eq:operators} we denoted with $'J_{S}$ the Jacobian of the transformation from the pre-interaction coordinates $('\mathbf{x},'\mathbf{x}_*)$ unto the post-interaction coordinates $(\mathbf{x},\mathbf{x}_*)$. Similarly for the pre-interaction feature coordinates, we denote as $'J_{F}$ the Jacobian of the transformation  $'c \mapsto c$. Finally, we denote as $'J_{B}$ the Jacobian of the transformation  $''c \to c$ associated with the binarization dynamics in \eqref{eq:binary-binarization}. The expectation $\left\langle \cdot\right\rangle$ in $Q_{S}(\cdot,\cdot)$ is taken with respect to the distribution of the random variables $\eta$ and $\eta_*$. A possible way to guarantee physical admissibility $\mathbf{x}',\mathbf{x}_*' \in \Omega$ and $c',c'' \in [0,1]^d$ has ben studied in \cite{CPT,TZ} for related dynamics where a cut-off function has been taken into account in the collision operator.

\subsection{Reduced Complexity Model: Fokker-Planck Limit}
While \eqref{eq:Boltzmann} provides a detailed mesoscopic description of the dynamics in terms of binary interactions, its analytical solution remains challenging due to the nonlinear and nonlocal structure of the collision operators. In order to derive a more tractable description, we consider a regime of a quasi-invariant interaction limit \cite{T,PT}, where each binary interaction produces only a small variation of the microscopic states, but interactions occur at a sufficiently high frequency.  In this limit, the cumulative effect of many small interactions leads to a nonlocal drift–diffusion equation of Fokker--Planck type for the distribution function $f(\mathbf{x},c,t)$.

We may rewrite \eqref{eq:Boltzmann} in weak form as follows

\begin{equation}    
\begin{aligned}
\frac{d}{dt}
&\iint_{\Omega\times[0,1]^d}
\psi(\mathbf{x},c)\,f(\mathbf{x},c,t)\,d\mathbf{x}\,dc =\theta_B \int_{\Omega\times[0,1]^d} (\psi(\mathbf{x},c'')-\psi(\mathbf{x},c))\,
f(\mathbf{x},c,t)\,d\mathbf{x}\,dc \,+\\
&\theta_F\iint_{\Omega\times[0,1]^d}
(\psi(\mathbf{x},c')-\psi(\mathbf{x},c))\,
f(\mathbf{x},c,t),d\mathbf{x}\,dc\,+ \\
&\theta_S\iint_{\Omega^2\times[0,1]^{2d}}
\left\langle\psi(\mathbf{x}',c)-\psi(\mathbf{x},c)
\right\rangle\,
f(\mathbf{x},c,t)\,f(\mathbf{x}_*,c_*,t)\,d\mathbf{x}\,dc\,d\mathbf{x}_*\,dc_*,
\end{aligned}
\label{eq:weak_boltzmann}
\end{equation}
where $\psi(\cdot,\cdot): \Omega \times [0,1]^d \to \mathbb{R}$ is a test function. We introduce a new time scale $\tau = \varepsilon t$ such that 
\begin{equation}
    f_{\varepsilon} := f_\varepsilon(\mathbf{x}, c, \tau/\varepsilon)\qquad f_{\varepsilon,*} := f_\varepsilon(\mathbf{x}_*, c_*, \tau/\varepsilon).
\end{equation}
Furthermore we scale the diffusion weight $\sigma^2(\mathbf{x},c)$ as follows $\sigma^2(\mathbf{x},c)\rightarrow \varepsilon\,\sigma^2(\mathbf{x},c)$. In the introduced time scale we observe that $\partial_\tau f_\varepsilon = \frac{1}{\varepsilon} \partial_t f_\varepsilon$.
Considering the regime where $\varepsilon\rightarrow 0^+$, and given that $\psi(\mathbf{x},c)$ is a sufficiently smooth test function, we notice that our scheme becomes quasi-invariant, which allows us to perform the following Taylor expansions
\begin{equation}
\label{eq:qi_scheme}
\begin{split}
\left\langle
\psi(\mathbf{x}',c) - \psi(\mathbf{x},c)
\right\rangle
&=
\left\langle \mathbf{x}'-\mathbf{x} \right\rangle
\cdot \nabla_{\mathbf{x}}\psi(\mathbf{x},c)
+ \frac{1}{2}
\left\langle (\mathbf{x}'-\mathbf{x})^{\!\top}
H_{\mathbf{x}}[\psi]
(\mathbf{x}'-\mathbf{x}) \right\rangle
+ R^S_\varepsilon[\psi](\mathbf{x},c). \\
\psi(\mathbf{x},c') - \psi(\mathbf{x},c)&=
(c'-c)\cdot\nabla_c \psi(\mathbf{x},c)
+ \frac{1}{2}(c'-c)^{\!\top}H_c[\psi](c'-c)
+ R^F_\varepsilon[\psi](\mathbf{x},c) \\
\psi(\mathbf{x},c'') - \psi(\mathbf{x},c)
&=
(c''-c)\cdot\nabla_c \psi(\mathbf{x},c)
+ \frac{1}{2}(c''-c)^{\!\top}H_c[\psi](c''-c)
+ R^B_\varepsilon[\psi](\mathbf{x},c).
\end{split}
\end{equation}
where $H_{\mathbf{x}}[\psi],H_c[\psi]$ are the Hessian matrices and $R^S_\varepsilon,R^F_\varepsilon,R^B_\varepsilon$ are remainder terms. Substituting the obtained Taylor expansion \eqref{eq:qi_scheme} into the weak form of the Boltzmann equation \eqref{eq:weak_boltzmann} we obtain

\begin{equation*}
\begin{split}
\frac{d}{d\tau}
&\int_{\Omega\times[0,1]^d}
\psi(\mathbf{x},c)\,f_\varepsilon(\mathbf{x},c,\tau)\,d\mathbf{x}\,dc
= \frac{\theta_S}{\varepsilon}
\int_{\Omega^2\times[0,1]^{d}}
\langle \mathbf{x}'-\mathbf{x} \rangle
\cdot \nabla_{\mathbf{x}}\psi(\mathbf{x},c)\,
f_\varepsilon f_{\varepsilon,*}\,
d\mathbf{x}\,dc\,d\mathbf{x}_*\,dc_* + 
\\
& + \frac{\theta_S}{2\varepsilon}
\int_{\Omega^2\times[0,1]^{2d}}
\Big\langle
(\mathbf{x}'-\mathbf{x})^{\!\top}
H_{\mathbf{x}}[\psi](\mathbf{x},c)
(\mathbf{x}'-\mathbf{x})
\Big\rangle
f_\varepsilon f_{\varepsilon,*}\,d\mathbf{x}\,dc\,d\mathbf{x}_*\,dc_* +
\\
&  \frac{\theta_F}{\varepsilon}
\int_{\Omega\times[0,1]^{d}}
( c'-c )\cdot\,
\nabla_c\psi(\mathbf{x},c)\,
f_\varepsilon \,d\mathbf{x}\,dc\,+ \\
&  \frac{\theta_F}{2\varepsilon}
\int_{\Omega\times[0,1]^{d}}
(c'-c)^{\!\top}H_c[\psi](\mathbf{x},c)(c'-c)\,
f_\varepsilon \,d\mathbf{x}\,dc\, + \\[4pt]
&  \frac{\theta_B}{\varepsilon}\int_{\Omega\times[0,1]^d} (c''-c)\cdot\nabla_c\psi(\mathbf{x},c)\,f_\varepsilon\,d\mathbf{x}\,dc + \\
&  \frac{\theta_B}{2\varepsilon}\int_{\Omega\times[0,1]^d} (c''-c)^{\!\top}H_{c}[\psi](\mathbf{x},c)(c''-c) f_\varepsilon\,d\mathbf{x}\,dc + \\
& \dfrac{1}{\varepsilon}\left( \int_{\Omega^2 \times[0,1]^{2d}} \theta_S R_\varepsilon^S[\psi](\mathbf{x},c) f_\varepsilon f_{\varepsilon,*} d\mathbf{x}d\mathbf{x}_*dc dc_* +\int_{\Omega \times[0,1]^d}(\theta_F R_\varepsilon^F[\psi](\mathbf{x},c) + \theta_B R_\varepsilon^B[\psi](\mathbf{x},c))f_\varepsilon) d\mathbf{x}dc\right)
\end{split}
\end{equation*}
Using standard arguments and provided $\eta,\eta_*$ have finite third order moment it is possible to prove that the remainder term 
\[
\begin{split}
\mathcal R_\varepsilon(f_\varepsilon,f_{\varepsilon,*}) = \dfrac{1}{\varepsilon}\left( \int_{\Omega^2 \times[0,1]^{2d}} \theta_S R_\varepsilon^S[\psi](\mathbf{x},c) f_\varepsilon f_{\varepsilon,*} d\mathbf{x}d\mathbf{x}_*dc dc_* +\right.\\
\left.\int_{\Omega \times[0,1]^d}(\theta_F R_\varepsilon^F[\psi](\mathbf{x},c) + \theta_B R_\varepsilon^B[\psi](\mathbf{x},c))f_\varepsilon) d\mathbf{x}dc\right)
\end{split}\]
is such that $|\mathcal R_\epsilon| \to 0$ as $\varepsilon\to 0^+$ in the limit where $\varepsilon\rightarrow 0^+$ \cite{T}. Therefore, in the regime $\varepsilon \to 0^+$ we have that $f_\varepsilon \to f$ solution to the following Fokker-Planck type equation
\begin{equation}\label{eq:complete_model}
\partial_t f(\mathbf{x},c,t)
+ \theta_F\nabla_c\cdot\!\left[(\alpha(\mathbf{x},t)-c)\phi(c)f(\mathbf{x},c,t)\right]
+ \theta_B\nabla_c \cdot\!\left(-\nabla_cV(c)f(\mathbf{x},c,t)\right)
= \theta_S\,\hat{Q}_S
\end{equation}
where
\begin{align}
\hat{Q}_S
&=
\nabla_{\mathbf{x}} \cdot
\left[
\mathcal{B}[f](\mathbf{x},c,t)\,f(\mathbf{x},c,t)
+
\frac{1}{2}\,\nabla_{\mathbf{x}}(\sigma^2(\mathbf{x},c)f(\mathbf{x},c,t))
\right].
\label{eq:space-operator}
\end{align}
and the operator $B[f](\mathbf{x},c,t)$ has the following form
\begin{equation}
B[f](\mathbf{x},c,t)=
\int_{[0,1]^d}\int_{\Omega}
P_{\Delta_2}(c,c_*) (\mathbf{x}-\mathbf{x}_*)\,f(\mathbf{x}_*,c_*,t)\,d\mathbf{x}_*\,dc_*.
\label{eq:nonlocal-operator}
\end{equation}
Furthermore, we notice that the obtained Fokker-Planck equation is endowed with no-flux boundary conditions in the spatial variable and in the feature variable, which read as follows

\begin{equation}
\begin{aligned}
\mathcal{B}[f](\mathbf{x},c,t)\,f(\mathbf{x},c,t) + \frac{1}{2}\,\nabla_{\mathbf{x}}(\sigma^2(\mathbf{x},c)f(\mathbf{x},c,t))\,\Big|_{\mathbf{x} = \partial\Omega } &= 0, \\
(\alpha(\mathbf{x},t)-c)\phi(c)f(\mathbf{x},c,t)\,\Big|_{c \in \partial[0,1]^d} &= 0, \\
-\nabla_cV(c)f(\mathbf{x},c,t)\,\Big|_{c \in \partial[0,1]^d} &= 0
\end{aligned}
\end{equation}

\subsection{Quasi-invariant Profiles}

In this section, we focus exclusively on the effects of the spatial operator $\hat{Q}_S$, which governs the spatial dynamics of the system in the absence of feature dynamics. As shown in Equation~\ref{eq:space-operator}, this operator combines a nonlocal drift term with a non-constant diffusion whose coefficient depends on both space and feature variables. Because of the nonlocal nature of the drift term, the stationary distribution cannot, in general, be written explicitly. Furthermore, we are interested in characterizing the preserved quantities of the operator $\hat{Q}_S$ given that, due to the structure of Equation~\ref{eq:complete_model}, we may obtain macroscopic equations for the evolution of such quantities. To do so, we start by considering a test function $\psi(\mathbf{x},c) = \psi(\mathbf{x})$ which only depends on the spatial variable and integrate in the spatial domain

\begin{equation}
    \frac{d}{dt} \int_{\Omega}\psi(\mathbf{x})\,f(\mathbf{x},c,t)\,d\mathbf{x} = \int_{\Omega} \psi(\mathbf{x})\nabla_{\mathbf{x}} \cdot \left( \mathcal{B}[f](\mathbf{x},c,t)\,f(\mathbf{x},c,t) + \frac{1}{2}\,\nabla_{\mathbf{x}}(\sigma^2(\mathbf{x},c)f(\mathbf{x},c,t))\right) d\mathbf{x}.
\end{equation}

By setting $\psi(\mathbf{x}) = 1$ we may observe that the macroscopic quantity

\begin{equation}
    \rho(c,t) = \int_{\Omega} f(\mathbf{x},c,t)\,d\mathbf{x}
\label{eq:rho}
\end{equation}

is preserved by the operator $\hat{Q}_S$ due to the no-flux boundary conditions. This quantity represents the fraction of particles with a given feature value $c$. In other words, it can be interpreted as the number of pixels with a given color intensity. Following the same procedure, we may consider $\psi(\mathbf{x}) = \mathbf{x}$ which represents

\begin{equation}
    m(c,t) = \int_{\Omega} \mathbf{x}\,f(\mathbf{x},c,t)\,d\mathbf{x}
\end{equation}

the mean spatial position of the particles with feature value $c$. By direct computation, it can be observed that such quantity is not preserved by the operator $\hat{Q}_S$. Furthemore, we may also consider the case where $\psi(\mathbf{x},c) = \mathbf{x}\,\chi(|c-c_*|<\Delta_2)$ which depends on both the spatial and feature variable. We write

\begin{equation}
\begin{aligned}    
    \frac{d}{dt}\int_{[0,1]^d}\int_{\Omega}\mathbf{x}\,\chi(|c-c_*|<\Delta_2)f(\mathbf{x},c,t)\,d\mathbf{x}\,dc &= \frac{d}{dt}\int_{[0,1]^d}\rho(c,t)\,m(c,t)\,\chi(|c-c_*|<\Delta_2)\,dc = \\
    = \int_{[0,1]^d}\int_{\Omega}\mathbf{x}\,\chi(|c-c_*|<\Delta_2)\,\nabla_{\mathbf{x}}\cdot\Big( B[f](\mathbf{x},c&,t)\,f(\mathbf{x},c,t) + \frac{1}{2}\,\nabla_{\mathbf{x}} (\sigma^2(\mathbf{x},c)f(\mathbf{x},c,t))\Big)\,d\mathbf{x}\,dc  = \\
    =  - \int_{[0,1]^{2d}} \chi^2(|c-c_*|<\Delta_2)\,\rho(c,t)\,\rho(&c_*,t)\,\big(m(c,t)-m(c_*,t)\big)\,dc\,dc_* = 0
\end{aligned}    
\end{equation}

Noticing that the last term is symetric with respect to the exchange between $c$ and $c_*$ we can conclude that the macrocopic quantity defined as

\begin{equation}
    \mathcal{F}(c,t) = \int_{[0,1]^d}\chi(|c-c_*|<\Delta_2)\,\rho(c_*,t)\,m(c_*,t)\,dc_*
    \label{eq:Fcant}
\end{equation}

is preserved by the operator $\hat{Q}_S$. Furthermore, we now that the quasi-equilibrium distribution associated with the operator $\hat{Q}_S$ is given by the stationary solution of the following equation

\begin{equation}
\mathcal{B}[f^{\infty}](\mathbf{x},c,t)\,f^{\infty}(\mathbf{x},c)
+ \frac{1}{2}\,\nabla_{\mathbf{x}} \cdot \big(\sigma^2(\mathbf{x},c)\,f^{\infty}(\mathbf{x},c)\big)
= 0.
\end{equation}

We recall that the form of the nonlocal operator $\mathcal{B}[f](\mathbf{x},c,t)$ is given by Equation~\eqref{eq:nonlocal-operator}. The stationary solution can be expressed in explicit form as

\begin{equation}
f^{\infty}(\mathbf{x},c)
=
C \exp\left\{
-2 \int
\frac{\mathcal{B}[f](\mathbf{x},c,t)}{\sigma^2(\mathbf{x},c)}
\, d\mathbf{x}
\right\},
\end{equation}

where $C$ is a normalization constant. To further simplify the expression of the quasi-equilibrium distribution, we rewrite the nonlocal operator $\mathcal{B}[f](\mathbf{x},c,t)$ as follows:

\begin{align}
\mathcal{B}[f](\mathbf{x},c,t)
&=
\mathbf{x}
\int_{[0,1]^d}
\chi\big(|c-c_*|<\Delta_2\big)\,\rho(c_*,t)\,dc_*
\nonumber\\
&\quad
-\int_{[0,1]^d}
\chi\big(|c-c_*|<\Delta_2\big)\,\rho(c_*,t)\,m(c_*,t)\,dc_*,
\end{align}

where we have performed the integration with respect to the spatial variable and expressed it in terms of the macroscopic moments $\rho(c,t)$ and $m(c,t)$. We notice that the second term of the non local operator is equivalent to the quantity $\mathcal{F}(c,t)$ defined above, hence we may write the following expression for the quasi-equilibrium distribution associated with the operator $\hat{Q}_SS$ as follows
\begin{equation}
f^{\infty}_{\mathcal{F},\rho}(\mathbf{x},c)
=
C \exp\left\{
-2 \int
\frac{1}{\sigma^2(\mathbf{x},c)} \left(\mathbf{x}\,
\displaystyle \int_{[0,1]^d}
\chi\big(|c-c_*|<\Delta_2\big)\,\rho(c_*,t)\,dc_*
-
\mathcal{F}(c,t)\right)
\, d\mathbf{x}
\right\}.
\label{eq:steady.state}
\end{equation}

We notice that our quasi-equilibrium distribution is parametrized by two macroscopic quantities $\rho(c,t)$ and $\mathcal{F}(c,t)$ which we have already proven to be preserved by the operator $\hat{Q}_S$. In particular, we notice that the asymptotic distribution of the spatial operator $\hat{Q}_S$ is characterized by a heterogeneous structure in the feature variable where we obtain a spatial structure depending on the feature level. The explicit characterization of the quasi-equilibrium distribution requires knowledge of the macroscopic quantities $\rho(c,t)$ and $\mathcal{F}(c,t)$. Therefore, we aim at obtaining a macroscopic equation that dictates the evolution of such quantities in the presence of the transport operators in the feature space of our complete model shown in Equation~\ref{eq:complete_model}.

\section{Derivation of the Macroscopic Model and Segmentation}\label{sect:macro}

According to the complete model shown in Equation~\ref{eq:complete_model} the evolution of our system is characterized by the combination of three different mechanisms. The spatial interactions determine the aggregation of pixels into heterogeneous spatial structures through the balance between nonlocal attraction and diffusion, while the feature variables evolve under the combined action of transport and binarization operators. In the regime where spatial interactions occur at a fast time scale, the dynamics of the system can be naturally described through the evolution of the macroscopic quantities in the feature space corresponding to the conserved quantities of the operator $\hat{Q}_S$. Within this framework, the transport dynamics in feature space promote the formation of homogeneous regions composed of pixels sharing similar feature values. Therefore, from an application perspective, the binarization process should act on a slower time scale in order to preserve the formation of homogeneous regions and correctly identify the different image segments. This motivates to consider the following scaling of the parameters $\theta_S,\theta_F,\theta_B$ 

\begin{equation*}
    \lim_{\theta_S \rightarrow +\infty} \frac{\theta_F}{\theta_S} = 0, \qquad \theta_F >> \theta_B.
\end{equation*}

\subsection{Multiscale Dynamics and Derivation}
In summary we can describe the dynamics of our system as follows: the spatial operator occurs at the fastest time scale, driving the system towards a quasi-equilibrium distribution in space. Then, operating at a time scale slower than the spatial dynamics but faster than the binarization process, the transport operator in the feature space generates homogeneous regions whule preserving the spatial structure of the quasi-equilibrium distribution. Finally, at the slowest rate, the binarization process pushes the entire mass toward the borderds of the feature space thus generating the final segmentation mask. In this regime we can perform a three-way splitting of the dynamics as follows

\begin{itemize}
  \item \textbf{Relaxation step:} $\partial_t f(\mathbf{x},c,t) = \hat{Q}_S$
  \item \textbf{Transport step:} $\partial_t f(\mathbf{x},c,t) +  \nabla_c \cdot \Big[(\alpha_{\Delta_1}(\mathbf{x},t)-c)\phi(c) f(\mathbf{x},c,t)\Big] = 0$ 
  \item \textbf{Binarization step:} $\partial_t f(\mathbf{x},c,t) +  \nabla_c \cdot \Big[-\nabla_cV(c) f(\mathbf{x},c,t)\Big] = 0$
\end{itemize}

In order to characterize the macroscopic evolution of $\rho(c,t)$ we consider a test function $\varphi(\mathbf{x})$ and write in weak form our complete model

\begin{equation}
\begin{aligned}
\frac{d}{dt}\Bigg(
\rho(c,t)\int_{\Omega}\varphi(\mathbf{x})\,f_{\mathcal F,\rho}^{\infty}(\mathbf{x},c)\,d\mathbf{x}
\Bigg)
&+ \theta_F\,\nabla_c\cdot\Bigg[
\phi(c)\,\rho(c,t)\int_{\Omega}\varphi(\mathbf{x})\big(\alpha_{\Delta_1}(\mathbf{x},t)-c\big)
\,f_{\mathcal F,\rho}^{\infty}(\mathbf{x},c)\,d\mathbf{x}
\Bigg]
\\
+ \nabla_c\cdot\Bigg[
-\nabla_cV(c)\,&\rho(c,t)\int_{\Omega}\varphi(\mathbf{x})\,f_{\mathcal F,\rho}^{\infty}(\mathbf{x},c)\,d\mathbf{x}
\Bigg]
= \int_{\Omega} \varphi(\mathbf{x}) Q_S d\mathbf{x}.
\end{aligned}
\end{equation}

We recall that the operator $\hat{Q}_S$ does not preserve the mean position of the pixels with a given feature thus we are not able to obtain a macroscopic equation for the evolution of $m(c,t)$. However, by setting $\varphi(\mathbf{x}) = 1$ we can obtain the evolution of $\rho(c,t)$ which we have already verified to be preserved by the operator $\hat{Q}_S$

\begin{equation}
    \partial_t \rho(c,t) + \nabla_c \cdot \left\{\theta_F \phi(c)\,\rho(c,t)\big(A(c,t)-c\big)- \nabla_cV(c)\,\rho(c,t)\right\}=0
\end{equation}

where $A(c,t):[0,1]^d \times \mathbb{R}_+ \to \mathbb{R}$ and $\alpha_{\Delta_1}(\mathbf{x},t):\mathbb{R}^2 \times \mathbb{R}_+ \to [0,1]^d$ are defined as follows

\begin{equation}
\begin{aligned}
A(c,t) &= \int_{\Omega}\alpha_{\Delta_1}(\mathbf{x},t)\,f_{\mathcal F,\rho}^{\infty}(\mathbf{x},c)\,d\mathbf{x} \\
\alpha_{\Delta_1}(\mathbf{x},t) &= \frac{\int_{\Omega\times[0,1]^d} c_* \chi\!\bigl(|\mathbf{x}-\mathbf{x}_*|<\Delta_1\bigr) f_{\mathcal F,\rho}^{\infty}(\mathbf{x}_*,c_*) dc_* d\mathbf{x}_*}{\int_{\Omega\times[0,1]^d} \chi\!\bigl(|\mathbf{x}-\mathbf{x}_*|<\Delta_1\bigr) f_{\mathcal F,\rho}^{\infty}(\mathbf{x}_*,c_*) dc_* d\mathbf{x}_*}.
\end{aligned}
\label{eq:A}
\end{equation}

We observe that the evolution of $\rho(c,t)$ depends on the quantity $A(c,t)$, which in turn depends on the quasi-equilibrium distribution $f^{\infty}(\mathbf{x},c)$. We recall that the expression of the quasi-equilibrium, given in Equation~\eqref{eq:steady.state}, is parametrized by the macroscopic quantities $\rho(c,t)$ and $\mathcal{F}(c,t)$. Therefore, in order to obtain a closed system, it is necessary to derive an additional equation governing the evolution of $\mathcal{F}(c,t)$. To this end, we proceed as before by selecting the test function
\[
\varphi(\mathbf{x},c)= \mathbf{x}\, \chi\big(|c-c_*|<\Delta_2\big),
\]
which leads to the following macroscopic system:

\begin{equation}
\left\{
\begin{aligned}
    \partial_t \rho(c,t) 
    &+ \nabla_c \cdot \Big(
    \theta_F \phi(c)\,\rho(c,t)\big(A(c,t)-c\big)\Big)    
    - \nabla_c \cdot \Big(
    \nabla_c V(c)\,\rho(c,t)\Big)
    = 0, 
    \\[6pt]
    \partial_t \mathcal{F}(c,t) 
    &+ \nabla_c \cdot \Big(
    2\,\Delta_2\,\phi(c)\,\rho(c,t)\,
    \big[
        E(c,t)
        - c \int_{\Omega}\mathbf{x}\,f_{\mathcal{F},\rho}^{\infty}(\mathbf{x},c)\,d\mathbf{x}
    \big]
    \Big)
    \nonumber\\
    \quad
    &- \nabla_c \cdot \Big(
    2\,\Delta_2\,\nabla_c V(c)\,\rho(c,t)
    \int_{\Omega}\mathbf{x}\,f_{\mathcal{F},\rho}^{\infty}(\mathbf{x},c)\,d\mathbf{x}
    \Big)
    = 0.
\end{aligned}
\right.
\label{eq:macro.model}
\end{equation}
Here, the quantity $E(c,t):[0,1]^d \times \mathbb{R}_+ \to \mathbb{R}^2$ is defined as
\begin{equation*}
    E(c,t) 
    =
    \int_{\Omega} \mathbf{x} \,\alpha_{\Delta_1}(\mathbf{x},t)\,
    f_{\mathcal{F},\rho}^{\infty}(\mathbf{x},c)\,d\mathbf{x}.
\end{equation*}

\subsection{Model-Oriented Macroscopic Segmentation Masks}

In this section, we outline the procedure used to obtain a segmentation mask from the macroscopic model. Given an input image, we define the feature values as
\begin{equation}
    c_{ij} = \frac{C_{ij} - \min_{i,j=1,\dots,N} C_{ij}}{\max_{i,j=1,\dots,N} C_{ij} - \min_{i,j=1,\dots,N} C_{ij}} \in [0,1],
\end{equation}
where $C_{ij}$ denotes the intensity of the pixel of the original image, $N$ is the total number of pixels, and $j$ indexes the channels used to represent the image. Each pixel is then associated with a spatial position and a feature vector, and we rescale the spatial domain to $\Omega = [-1,1] \times [-1,1]$. Using a histogram-based approach, we introduce $N_x$, $N_y$, and $N_c$ as the number of bins used to reconstruct the initial distribution $f_0(\mathbf{x},c)$. From Equations~\eqref{eq:rho} and~\eqref{eq:Fcant}, we estimate the initial feature distribution $\rho(c,0)$ and the macroscopic quantity $\mathcal{F}(c,0)$. The asymptotic distribution is obtained by computing the evolution of the macroscopic system \eqref{eq:macro.model} using an explicit Euler discretization in time, combined with a Rusanov-reconstruction for the numerical fluxes. We emphasize that the fluxes exhibit an explicit dependence on the feature variable $c$, resulting in a local behavior that differs from standard hydrodynamic fluxes. Furthermore, due to the non-conservation of the mean pixel position the resulting macroscopic system is a first order system. In this way, given  an image and its segmented region, we can use our macroscopic model so as to reproduce the region of interest (ROI). We seek to determine the optimal parameters for which the asymptotic feature distribution $\rho^{\infty}(c)$ reproduces the distribution of the Ground Truth Segmentation Mask (GTSM), denoted by $\rho_{\mathrm{GTSM}}(c)$. This naturally leads to the optimization problem

\begin{equation}
    \min_{\Delta_1,\Delta_2,\sigma^2,\tilde{c}} 
    \left\|\rho^{\infty}(c) - \rho_{\text{GTSM}}(c)\right\|.
\label{eq:opt}
\end{equation}

A key advantage of this reduced macroscopic formulation is its significantly lower computational cost compared to previously introduced consensus-based segmentation models. This reduction enables the use of more efficient optimization strategies that would otherwise be computationally prohibitive due to the high cost of evaluating the segmentation masks at a high number of iterations. Moreover, this framework allows for the exploration of different discrepancy measures between distributions, opening the door to future investigations on how the choice of distance metric influences the resulting segmentation. Once the optimal parameters are obtained, we compute the long-time solution of the particle system following the approach of the previous work~\cite{CTZ,CPLFZ}. In this way we generate the segmented image using optimal set of parameters obtained from the macroscopic model. 

\section{Numerical Results}\label{sect:num}

In this section, we report several numerical experiments to illustrate the consistency of the proposed model and to highlight its applicability to the segmentation of images. 

First, we will concentrate on the long-time solution of the Boltzmann-type equation for the spatial dynamics without considering any dynamics in the feature space. In particular, we will show the consistency between the long-time solution of the Boltzmann-type operator and the Fokker-Planck formulation (\ref{eq:space-operator}) in the quasi-invariant regime ($\varepsilon << 1$). To do so, we will make use of the well studied Direct Simulation Monte Carlo (DSMC) method to approximate the long-time solution of the Boltzmann-type equation. For the interest readers we refer to \cite{DP,PR,PT} for a detailed description of the method.

Secondly, we will provide numerical evidence of the consistency between the macroscopic model given by Equation~\ref{eq:macro.model} with the complete Boltzmann-type equation, where we include the transport operator, in the regime where the spatial interactions occur at a fast rate $\theta_S \to \infty$ . In these consistency tests we will consider a one-dimensional feature space.

Finally, we will focus on the application of the complete model, given by Equation~\ref{eq:complete_model}, to the segmentation of different geometric shapes with a noise background. We will showcase the optimization procedure to find the optimal parameters and we will consider different choices of image noises to show the robustness of the model to different noise configurations. To obtain the long time solution of Equations~\ref{eq:macro.model} we will make use of Rusanov-type finite volume scheme where a detailed description of the method can be found in \cite{Toro,TT}.

\subsection{Test 1: consistency of the reduced complexity model}

In this test we showcase the consistency between the Boltzmann-type formulation given by \ref{eq:Boltzmann}, in the absence of the binarization operator, and the Fokker-Planck Equation~\ref{eq:space-operator} in the quasi-invariant regime considering a one dimensional feature space. In this first test, we set the transport function $\phi(c) = 0$ such that we only consider the spatial dynamics of the system. Under these assumptions we recall that the asymptotic distribution given by the Fokker-Planck operator is given by \ref{eq:steady.state}.  We start by writing the Boltzmann-type equation in strong form

\begin{equation}
    \partial_t f(\mathbf{x},c,t) = \frac{1}{\varepsilon}(Q^+_{S}(f,f)(\mathbf{x},c,t) - f(\mathbf{x},c,t))
\end{equation}
where 
\begin{equation}
Q^+_{S}(f,f)(\mathbf{x},c,t) = \int_{\Omega}\int_0^1 \left\langle \frac{1}{'J_{S}}f('\mathbf{x},c,t)f('\mathbf{x}_*,c_*,t)  dc_* d\mathbf{x}_* \right\rangle
\end{equation}
where $'J_{S}$ is the Jacobian of the transformation from pre-interaction coordinates $('\mathbf{x},c,'\mathbf{x}_*,c_*)$ unto the post-interaction variables $(\mathbf{x},c,\mathbf{x}_*,c_*)$. With the choice of $\phi(c) = 0$ the dynamics in the feature space remain static. 
In order to compute the solution we adopt a DSMC scheme based on the Nanbu algorithm for Maxwellian molecules \cite{PT}. We discretise the time domain as $t^n = n\Delta t$ with a fix time step $\Delta t > 0$. We denote as $f^n(\mathbf{x},c)$ the approximation of the distribution function at time $t^n$. A forward discretization allows us to write
\begin{equation}
f^{n+1}(\mathbf{x},c) = \left(1-\frac{\Delta t}{\varepsilon}\right)f^n(\mathbf{x},c) + \frac{\Delta t}{\varepsilon} Q^+_{S}(f^n,f^n)(\mathbf{x},c)
\end{equation}
Provided $\Delta t < \varepsilon$, we can interpret the previous equation as a convex combination of the distribution function at time $t^n$ and the gain term of the Boltzmann-type operator. This formulation implies that, within a time step $\Delta t$, a pair of particles interacts with probability $\Delta t / \varepsilon$, while with probability $1 - \Delta t / \varepsilon$ no interaction occurs. In this first test, we consider $N = 10^5$ particles, $\varepsilon = 5\times10^{-1},10^{-1},10^{-2}$ and final time $T = 50$, $\Delta_1 = 2$, $\Delta_2 = 1$ and $\sigma^2 = 0.01$ in \eqref{eq:binary-interaction}.

\begin{figure}
\centering
\includegraphics[scale = 0.4]{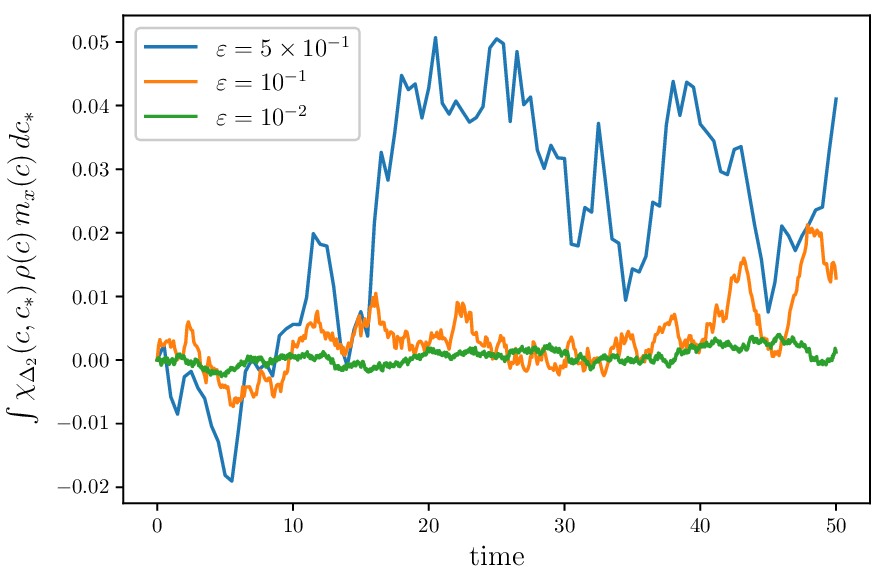}
\includegraphics[scale = 0.4]{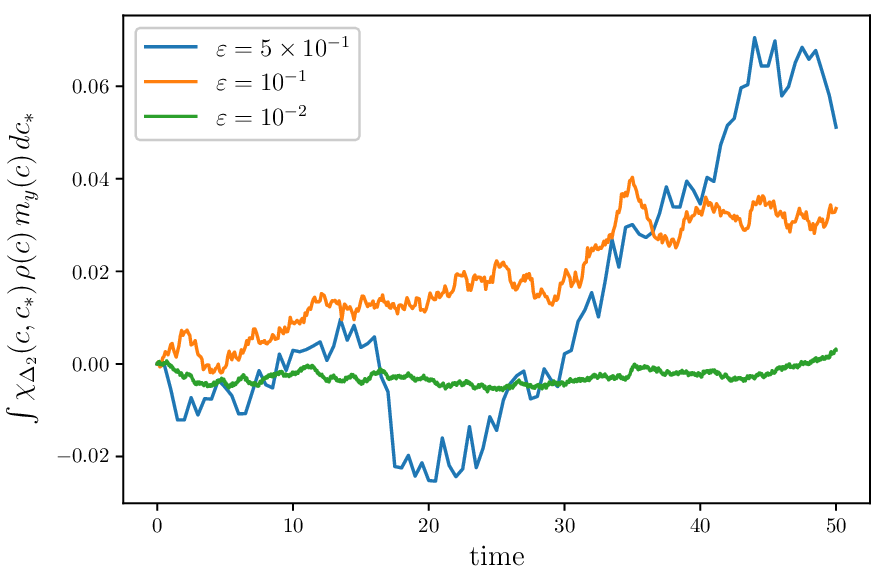}
\caption{\textbf{Test 1}. Time evolution of $\mathcal{F}(c,t) = \int_{\mathbb{R}^2} \mathbf{x} f(\mathbf{x},c,t)\,d\mathbf{x}$ with $\varepsilon$ = $10^{-2},10^{-1},5\times10^{-1}$. It can be observed that as, for small values of $\varepsilon> 0$, the quantity $\mathcal{F}(c,t)$ approaches to a constant value in time as expected given that it is a conserved quantity of the operator $\hat{Q}_{S}$.}
\label{fig:Fcant}
\end{figure}

\begin{figure}
\centering
\includegraphics[scale = 0.35]{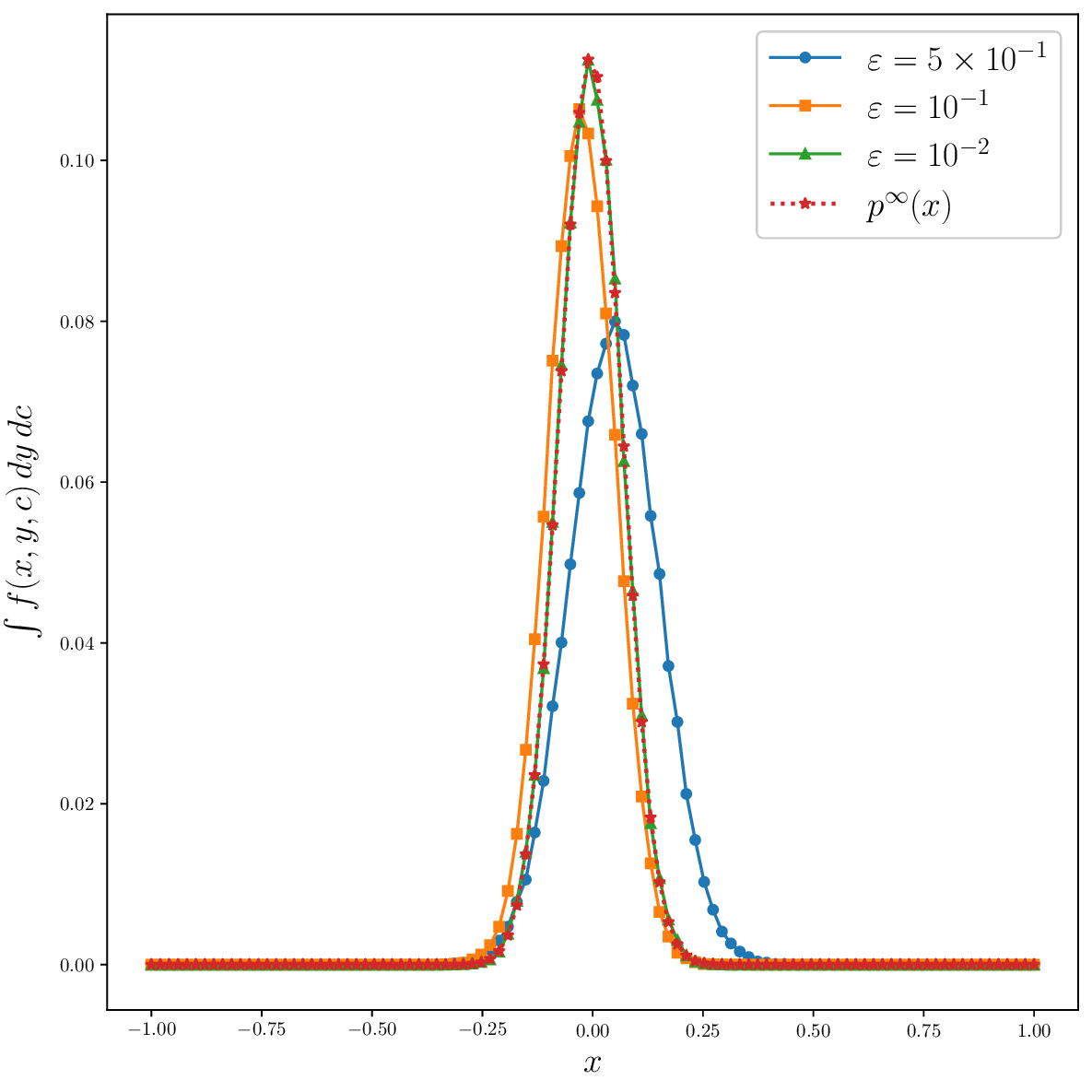}
\includegraphics[scale = 0.35]{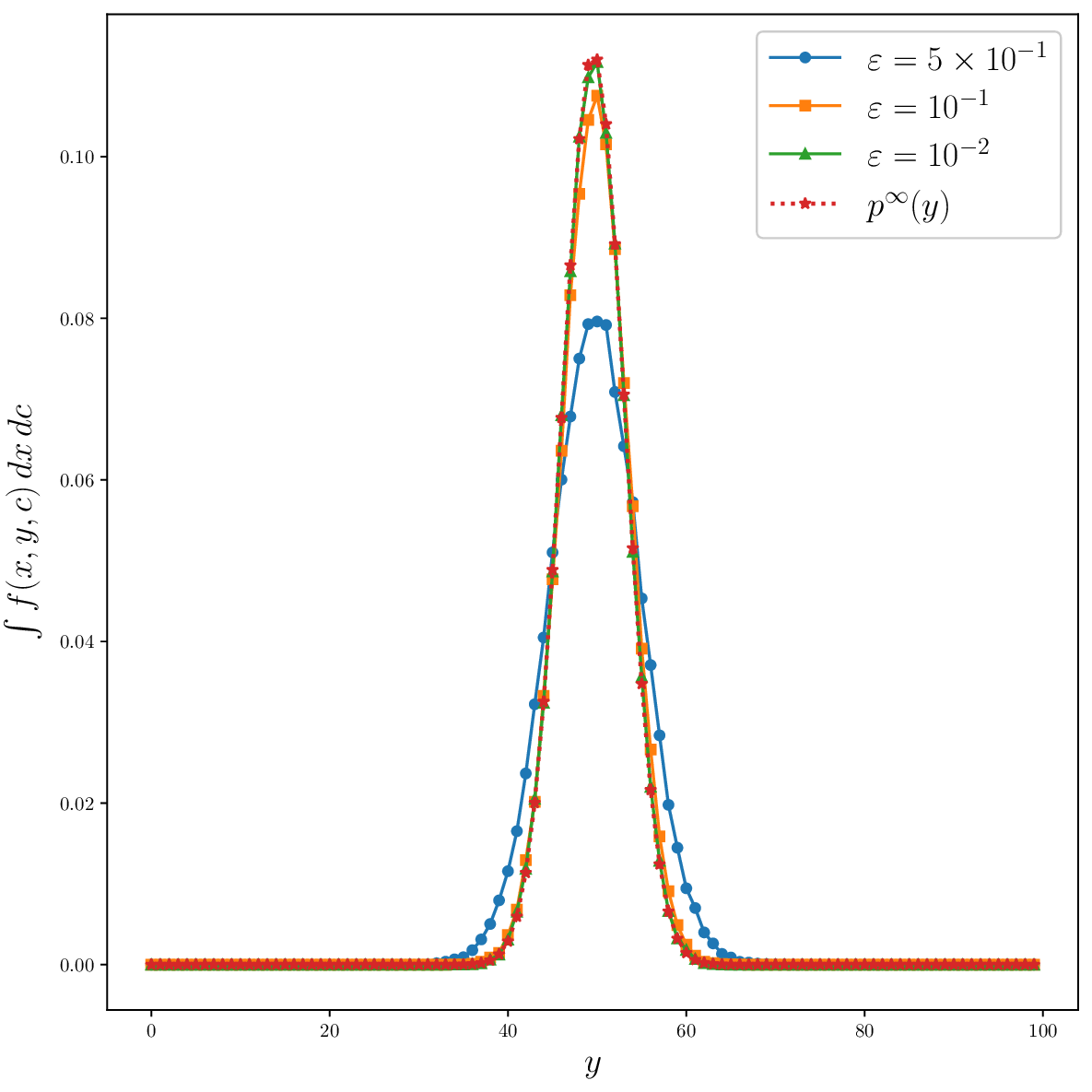}
\caption{\textbf{Test 1}. Comparison between the marginal distributions of the quasi-equilibrium distribution $f^{\infty}_{\mathcal{F},\rho}(\mathbf{x},c)$ given by Equation~\ref{eq:steady.state} and $f^{\infty}_{\epsilon,N}(\mathbf{x},c)$ obtained with the DSMC method with a final time $T = 50$ and $N = 10^5$ particles for different values of $c$ with $\varepsilon = 5 \times 10^{-1}, 10^{-2}$. We observe how the marginal distribution of the particle system converges to the marginal distribution of the quasi-equilibrium distribution for small values of $\varepsilon>0$.}
\label{fig:slices}

\end{figure}

We consider a spatially symmetric initial distribution and a Gaussian feature distribution $c \sim \mathcal{N}(0.5, 0.01)$. In Figure~\ref{fig:Fcant} we compute the quantity $\mathcal{F}(c,t)$ defined in Equation~\ref{eq:Fcant} for different values of $\varepsilon$. We recall that it was observed in the previous section that the Fokker-Planck operator preserves such quantity. We notice that for the chosen value of $\Delta_2$ big enough, such that all the particles interact between each other, the quantity $\mathcal{F}(c,t)$ looses its dependence on $c$. As we lower the value of $\varepsilon$ the fluctuations associated with the Monte Carlo scheme are reduced and it converges to a constant value in time.
Next, we compare the long-time distribution of the marginals obtained through the Boltzmann-type operator and the quasi-equilibrium distribution of the Fokker-Planck operator given by~\ref{eq:steady.state}. For the density reconstruction from the particle system, we approximate the asymptotic density by introducing a mollifier of the atomic measure
\begin{equation}
f_{\epsilon,N}(\mathbf{x},c) = \frac{1}{N}\sum_{i=1}^N \psi_{\sigma_\epsilon^2}(|(\mathbf{x},c)-(\mathbf{x}_i,c_i)|)
\end{equation}
with $\psi_{\sigma_\epsilon^2}$ a suitable smooth function. A classical choice is the histrogram case, that is obtained by setting $\psi_{\sigma_\epsilon^2}(r) = \chi(|r| \le \sigma_\epsilon^2/2)$ with $\sigma_\epsilon^2 = \Delta x$, being $\chi(\cdot)$ the indicator function. The results are shown in Figure~\ref{fig:slices}. We highlight how the marginal distributions of the reconstructed density converge to the quasi-equilibrium distribution as $\varepsilon \to 0$. Furthermore, we compute the first-order moment of both distributions, defined as
\begin{equation}
\rho(c,t)m(c,t) = \int_{\Omega} \mathbf{x} f(\mathbf{x},c,t)\,d\mathbf{x}.
\end{equation}

\begin{figure}
\centering
\begin{subfigure}{0.48\textwidth}
\centering
\includegraphics[width=\textwidth]{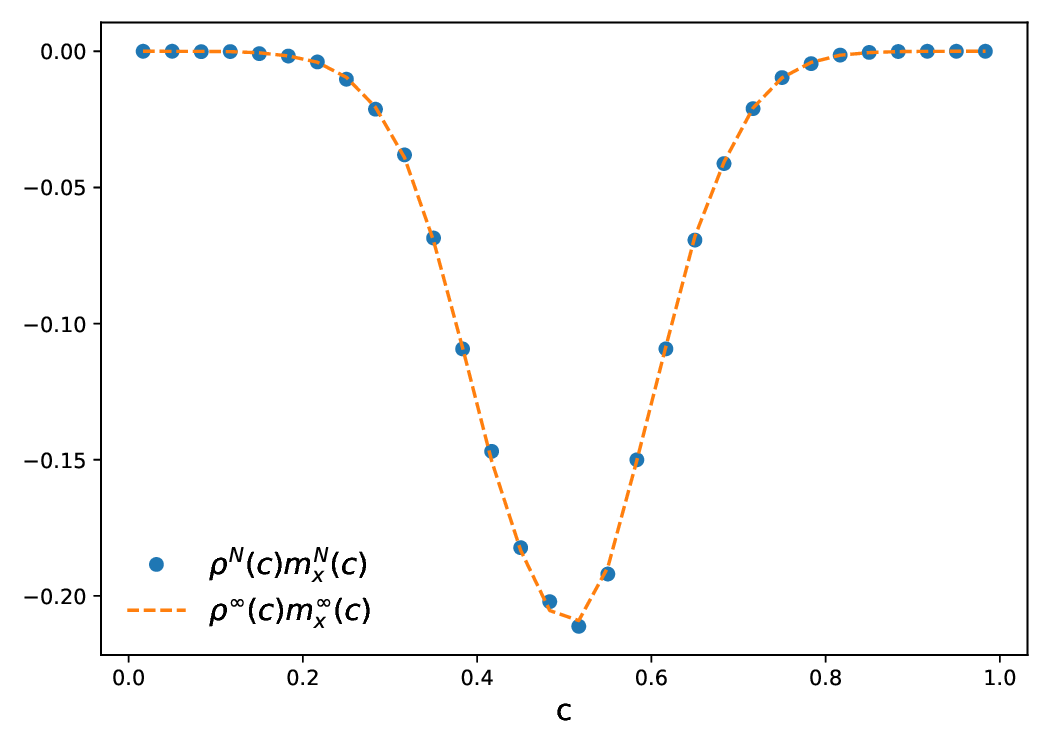}
\caption{}
\end{subfigure}
\hspace{0.2cm}
\begin{subfigure}{0.48\textwidth}
\centering
\includegraphics[width=\textwidth]{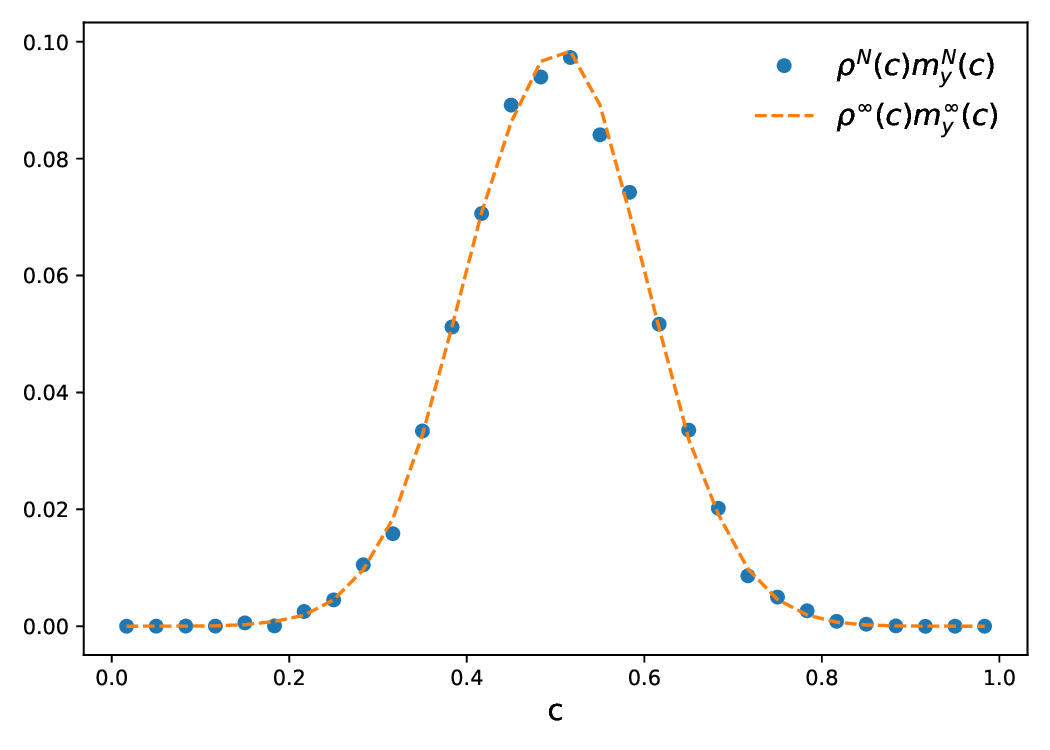}
\caption{}
\end{subfigure}
\caption{\textbf{Test 2}. We compute the first order moment of the distribution $\rho(c,t)m(c,t) = \int_{\mathbb{R}^2} \mathbf{x} f(\mathbf{x},c,t)\,d\mathbf{x}$ with $\varepsilon = 10^{-2}$ as computed from the particle distribution $f^{\infty}_{\epsilon,N}(\mathbf{x},c)$ and the quasi-equilibrium distribution $f^{\infty}_{\mathcal{F},\rho}(\mathbf{x},c)$ shown in Equation~\ref{eq:steady.state}. We recall that this quantity is not a conserved quantity of the system and we observed a good agreement between both quantities for $\varepsilon = 10^{-2}$ and final time $T = 50$. In the test we consider $N = 10^5$ particles.}
\label{fig:rhom}
\end{figure}

We recall that this quantity is not conserved by the spatial operator. The equilibrium distribution is parametrized by the macroscopic quantities $\mathcal{F}(c,t)$ and $\rho(c,t)$, which are evaluated only at the initial time, since both are preserved by the Fokker–Planck operator. The results are shown in Figure~\ref{fig:rhom}, where we set $\varepsilon = 10^{-2}$. As expected, in the quasi-invariant regime, the first-order moment computed from the particle system reproduces that of the quasi-equilibrium distribution.

\subsection{Test 2: consistency of the macroscopic model}

In this following test we want to show the consistency between our Boltzmann model and the macroscopic model, presented in Section~\ref{sect:macro}, in the absence of the binarization dynamics again in a one dimensional setting. In this case we have that the Boltzmann-type equation reads as follows
\begin{equation}
    \partial_t f(\mathbf{x},c,t) = \theta_S\, Q_S(f,f)(\mathbf{x},c,t) + \theta_F\,Q_F(f)(\mathbf{x},c,t)
\end{equation}
where we set $\theta_S = \frac{1}{\tau}$ and $\theta_F = 1$ to control the different time scales at which the spatial and feature dynamics evolve. Based on the approach shown in \cite{Z}, we approximate the solution of the Boltzmann-type equation following a time-splitting strategy where we first solve the spatial interaction dynamics $f^\ast=\mathcal{I}_{\Delta t}(f^\ast,f^\ast)$

\begin{equation}
\begin{split}
    &\partial_t f^\ast(\mathbf{x},c,t) = \theta_S Q_S(f^\ast,f^\ast)(\mathbf{x},c,t) \\
    &f^*(\mathbf{x},c,0)= f^n(\mathbf{x},c),
\end{split}
\end{equation}
and then applied the transport in the feature space $f^{\ast\ast} = \mathcal{T}_{\Delta t}(f^\ast)$
\begin{equation}
\begin{split}
    &\partial_t f^{\ast\ast}(\mathbf{x},c,t) + \theta_F \partial_c \left( [A(c,t) - c] f^{\ast\ast}(\mathbf{x},c,t) \right) = 0 \\
    &f^{\ast\ast}(\mathbf{x},c,0) = f^\ast(\mathbf{x},c,\Delta t)
\end{split}
\end{equation}
Hence, the solution at time $t^{n+1}$ is given by the combination of the two described steps as 
\[
f^{n+t}(\mathbf{x},c) = \mathcal T_{\Delta t}(\mathcal I_{\Delta t}(f^n,f^n)).
\]
In order to approximate the asymptotic solution of the macroscopic model given by Equation~\ref{eq:macro.model} we consider a finite volume discretization. We discretize our domain of size $L_x \times L_y \times L_c$ into a grid of size $N_x \times N_y \times N_c$ with mesh spacing $\Delta x = \frac{L_x}{N_x}$, $\Delta y = \frac{L_y}{N_y}$, and $\Delta c = \frac{L_c}{N_c}$. We consider the following initial condition 
\begin{equation}
    \rho(c,0) = \frac{1}{2}\chi(c \in [0,1])
\end{equation}
where we set $L_x = L_y = 2$ and $L_c = 1$ with $N_x = N_y = N_c = 30$. We introduce a time step $\Delta \hat{\tau}$ such that the following CFL condition is satisfied 
\begin{equation}
    \frac{\Delta \hat{\tau}}{\Delta c} \phi(c) \max_{c \in [0,1]} \left|[A(c,t)-c]\right| \leq 1    
\end{equation}
where the quantity $A(c,t)$ is defined in Equation~\ref{eq:A}. The system is solved by considering an Euler discretization in time and a Rusanov-type reconstruction of the numerical flux \cite{TT}. In the following, we consider $N = 10^5$ particles, $\tau = 1,10^{-3}$, $\Delta_1 = 0.2$, $\Delta_2 = 0.5$ and $\sigma^2 = 0.01$. 

\begin{figure}
    \centering
    \includegraphics[width=\textwidth]{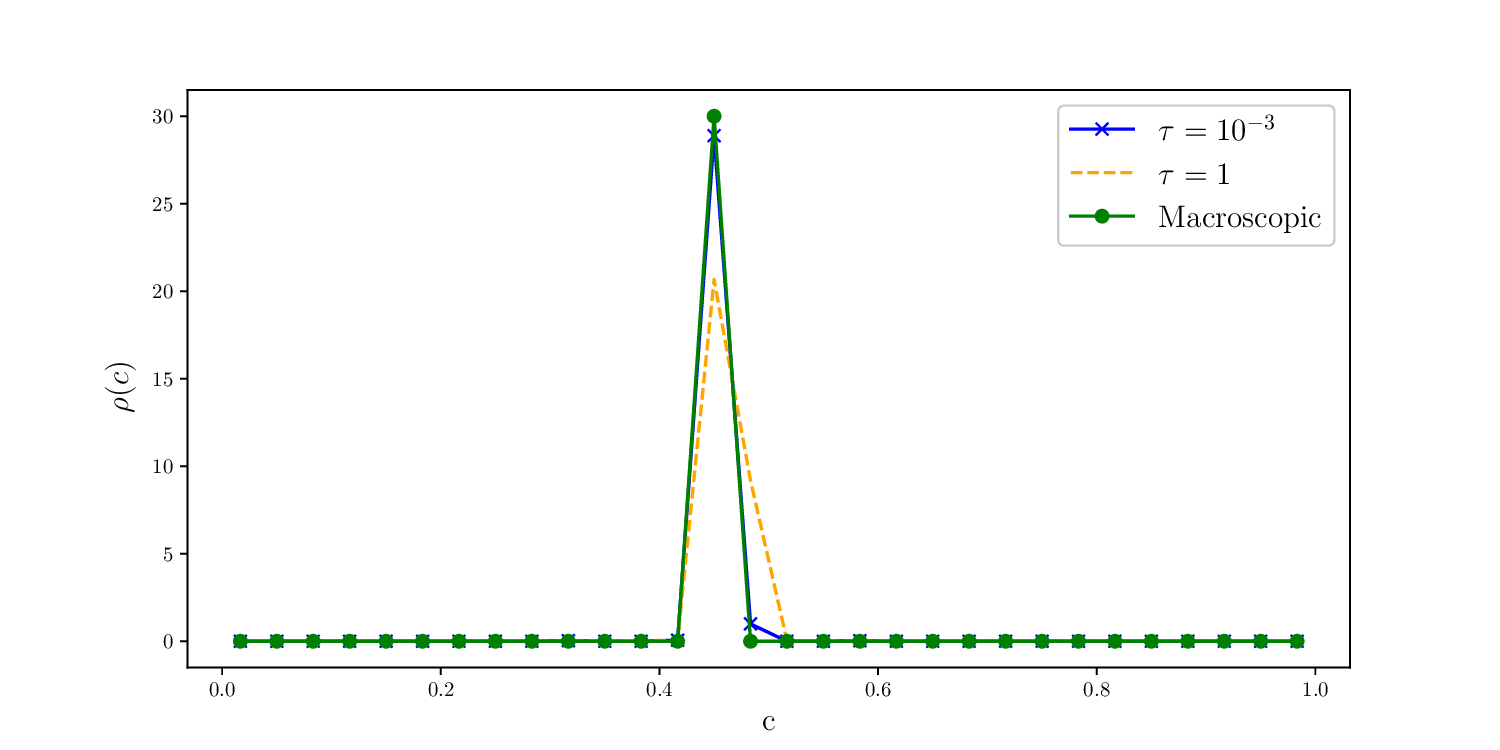}
    \caption{\textbf{Test 2}. We show the large-time  distribution $\rho^{\infty}(c)$ obtained with the Boltzmann-type equation for $\tau = 1$ and $\tau = 10^{-3}$ compared to the asymptotic distribution of the macroscopic model obtained evolving Equation~\ref{eq:macro.model}. We observe that as $\tau \rightarrow 0^+$ the solution of the Boltzmann-type equation approaches to the solution of the macroscopic model.}
    \label{fig:macro}
\end{figure}

The results are shown in Figure~\ref{fig:macro}. It can be seen that as $\tau$ decreases the solution of the Boltzmann-type equation converges to the solution of the macroscopic model. In this setting, the transport operator drives the feature variables toward a unique equilibrium value.

\subsection{Test 3: segmentation of geometric shapes through the macroscopic model}\label{sect:test3}

In this section we showcase the application of the complete macroscopic model to the segmentation of grayscale images in the presence of a noise background. We define an image as an intensity field $I(x,y)$ where $(x,y)$ represent the coordinates of a pixel. The noise field of an image is symbolized by $\tilde{I}(x,t) = \mathcal{G}(I)$ where $\mathcal{G}$ represent the different noise types that may corrupt an image. In this section we considered the following noise structure
\begin{equation}
\tilde{I}(x,y) = \begin{cases}
\mathcal{G}_G(I),\quad \text{if } (x,y) \in I_G \\
\mathcal{G}_B(I),\quad \text{if } (x,y) \in I_B
\end{cases}
\end{equation}
where $I_G$ and $I_B$ represent the regions of the image corresponding to the geometric shape and the background, respectively. Thus we are able to consider different noise types for the shape and the background with different noise intensities. In this way we can test the robustness of our method under various noise conditions. In the following tests we consider several noise structures. Gaussian noise is modeled by a zero-mean normal random variable

\begin{equation}\label{eq:gaussian_noise}
    n_\ell \sim \mathcal N(0,\sigma_\ell^2),\qquad \ell \in \{G,B\} 
\end{equation}
where we indicated with $G,B$ the different regions of the image. The probability density function of the Gaussian noise is given by
\begin{equation}
    f_{\mathcal{N},\ell(x)}
    =
    \frac{1}{\sqrt{2\pi\sigma_\ell^2}}
    \exp\!\left(
        -\frac{x^2}{2\sigma_\ell^2}
    \right).
\end{equation}
The corresponding noisy image is obtained through additive perturbation,
\begin{equation}
    \tilde I(x,y)
    =
    I(x,y)+n_\ell.
\end{equation}
We also consider uniform noise, defined by the probability density
\begin{equation}\label{eq:uniform_noise}
    f_{\mathcal{U},\ell(x)}(x)
    =
    \frac{1}{2u_\ell}
    \chi_{[-u_\ell,u_\ell]}(x),\qquad \ell \in \{G,B\}
\end{equation}
where \(\chi_A\) denotes the indicator function of the set \(A\). The noisy image is then given by
\begin{equation}
    \tilde I(x,y)
    =
    I(x,y)+n_{\ell},
    \qquad
    n_{\ell} \sim \mathcal U(-u_\ell,u_\ell).
\end{equation}
Furthermore, we consider multiplicative speckle noise where the noisy image is generated according to
\begin{equation}\label{eq:speckle_noise}
    \tilde I(x,y)
    =
    I(x,y)\bigl(1+n_{\ell}\bigr).
\end{equation}
In this case, we may observe that the variance of the perturbation increases proportionally to the square of the pixel intensity, which is characteristic of multiplicative noise. Finally, we consider Poisson noise. A Poisson random variable \(K\) with parameter \(\lambda\) is distributed according to
\begin{equation}
    \mathbb P(K=k)
    =
    \frac{\lambda^k e^{-\lambda}}{k!},
    \qquad
    k=0,1,2,\ldots.
\end{equation}
In our implementation, a pixel value \(I(x,y)\) is transformed according to
\begin{equation}\label{eq:poisson_noise}
    K(x,y)
    \sim
    \operatorname{Poisson}
    \!\left(
        \frac{I(x,y)}{\mu_\ell}
    \right),
\end{equation}
and the noisy image is given by
\begin{equation}
    \tilde I(x,y)
    =
    \mu_\ell\,K(x,y),
\end{equation}
where \(\mu_\ell>0\) is a scaling parameter controlling the noise level. For a more extensive presentation of the different noise types we point the reader to \cite{BJ}. 
The objective is to solve the optimization problem given by Equation~\ref{eq:opt} to find the optimal parameters of the model such that the large-time distribution $\rho(c,T)$, with T sufficiently large, approximates the distribution of the Ground Truth Segmentation Mask (GTSM) $\rho_{\text{GTSM}}(c)$. In this work we consider the $L^1$-norm as a measure of the discrepancy between the two distributions which is given by
\begin{equation}
    \left\|\rho(c,T)-\rho_{\text{GTSM}}(c)\right\|_1 = \frac{\int_0^1 \left|\rho(c,T)-\rho_{\text{GTSM}}(c)\right| dc}{\int_0^1 \rho_{\text{GTSM}}(c) dc}.
\end{equation}
From a computational perspective, we carry out the optimization procedure using the Consensus-Based Optimization (CBO) method. A particle-based method where a number of agents explore the landscape of the objective function at their given positions and define a consensus point, $c_\alpha$, which is an approximation of the global minimizer $x^*$. The consensus point is constructed by weighting the position of each agent against a Gibbs-type distribution
\begin{equation}
c_\alpha = \frac{\sum_{i=1}^{N_p} c_i \exp(-\alpha f(c_i))}{\sum_{i=1}^{N_p} \exp(-\alpha f(c_i))}
\end{equation}
for some constant $\alpha > 0$. In this way, we are able to give more weight to the agents with lower objective function values. Once this point is defined the positions of the agents evolve according to the following system of SDEs
\begin{equation}
dx_i^t = -\lambda (x_i^t - c_\alpha) dt + \sigma_{\text{CBO}} |x_i^t - c_\alpha| dW_t^i  
\end{equation}
where $\lambda > 0$ and $\sigma_{\text{CBO}} > 0$ are positive model parameters, and $\{W^i\}_{i=1}^N$ denotes a standard set of independent Wiener processes. In this formulation, the drift term drives the agents toward the consensus point, while the scaled diffusion promotes exploration of the objective function landscape, preventing convergence to local minima. For the interested readersm we refer to the following work for a detailed description of this method and its possible variations \cite{BBGRTV,BWR,CBE,CJLZ}. In all the tests we consider the following settings for the optimizer

\begin{itemize}
    \item Numer of particles: $N_p = 64$
    \item Number of iterations: $N_{\text{iter}} = 640$
    \item $\lambda = 1.0$, $\sigma^2_{\text{CBO}} = 0.5$ and $\alpha =12$
\end{itemize}

At each iteration of the CBO, we compute the asymptotic macroscopic quantities using the same numerical scheme introduced in the previous section, where we consider an Euler discretization in time and a Rusanov-type numerical flux. For all the tests, we consider $\tau_2 = 0.1$ and $\tau_1 = 10^{-3}$ such that we approximate the macroscopic regime. All the images considered in this section are of size $40\times 40$ pixels. After the optimal parameters were found, we obtained the segmentation mask using the DSMC method for the complete model. The details are shown in Algorithm~\ref{alg:DSMC_complete}.

\begin{algorithm}
\caption{Asymptotic particle-based algorithm for the approximation of the complete model~\ref{eq:complete_model}.}
\begin{algorithmic}[1]

\State Fix $\tilde{\varepsilon} = \frac{\tau_1 \tau_2}{\tau_1 + \tau_2}$ and $N_s$
\State Sample $\{(\mathbf{x}_i^n, c_i^n)\}_{i=1}^{N_s}$ from $f^n$

\State Set 
\[
N_1 = \frac{\tilde{\varepsilon}}{\tau_1} N_s, \quad
N_2 = \frac{\tilde{\varepsilon}}{\tau_2} N_s, \quad
N_3 = \tilde{\varepsilon} N_s
\]

\vspace{0.2cm}

\For{$k = 0:\Delta t$ with step $\tilde{\varepsilon}$}
    \State $N_c = \mathrm{round}(N_1/2)$
    \State Select $N_c$ random pairs $(i,i_*)$

    \For{$i = 1:N_c$}
        \State Sample $\eta_i, \eta_{i_*} \sim \mathcal{N}(0, I)$

        \State Update space positions:
        \[
        \begin{cases}
        \mathbf{x}_i^{n+1} = \mathbf{x}_i^n 
        + \tilde{\varepsilon} P_{\Delta_2}(c_i^n, c_{i_*}^n)(\mathbf{x}_{i_*}^n - \mathbf{x}_i^n)
        + \sqrt{2\sigma^2 \tilde{\varepsilon}} \, \eta_i, \\[0.2cm]
        \mathbf{x}_{i_*}^{n+1} = \mathbf{x}_{i_*}^n 
        + \tilde{\varepsilon} P_{\Delta_2}(c_{i_*}^n, c_i^n)(\mathbf{x}_i^n - \mathbf{x}_{i_*}^n)
        + \sqrt{2\sigma^2 \tilde{\varepsilon}} \, \eta_{i_*}
        \end{cases}
        \]
    \EndFor
\EndFor

\vspace{0.2cm}

\For{$k = 0:\Delta t$ with step $\tilde{\varepsilon}$}
    \For{$i = 1:N_2$}
        \State Transport feature values:
        \[
        c_i^{n+1} = c_i^n 
        - \tilde{\varepsilon} (\alpha(\mathbf{x}_i^n,t) - c_i^n)\phi(c_i^n)
        \]
    \EndFor
\EndFor

\vspace{0.2cm}

\For{$k = 0:\Delta t$ with step $\tilde{\varepsilon}$}
    \For{$i = 1:N_3$}
        \State Binarization step:
        \[
        c_i^{n+1} = c_i^n 
        - \tilde{\varepsilon}\, \nabla_{c} V_\varepsilon(c_i^n)
        \]
    \EndFor
\EndFor

\end{algorithmic}
\label{alg:DSMC_complete}
\end{algorithm}

\subsubsection{Test 3.1: impact of different shapes on the segmentation method}

In this section, we perform various tests for different geometric shapes centered on a noisy background under various noise configurations. Throughout all experiments, the foreground and background are corrupted by the same noise type, although a lower noise intensity is prescribed for the background region. 

\begin{table}
\centering
\renewcommand{\arraystretch}{1.3}

\begin{tabular}{|c|c|c|c|c|c|}
\hline
 & $\Delta_1$ & $\Delta_2$ & $\sigma^2$ & $\tilde{c}$ & Loss \\
\hline
Square   & 0.2903 & 0.4685 & 0.1549 & 0.4778 & 0.0226\\
\hline
Circle   & 0.2138 & 0.1537 & 0.1350 & 0.5140 & 0.0249\\
\hline
Triangle & 0.1108 & 0.1657 & 0.0778 & 0.5287 & 0.0179\\
\hline
Rhombus  & 0.1720 & 0.1564 & 0.0844 & 0.5112 & 0.0137\\
\hline
\end{tabular}
\caption{\textbf{Test 3.1}. Optimal parameters for the different geometric shapes with varying noise configurations from loss minimization \eqref{eq:opt}. {Square}: Gaussian noise with $\sigma^2_G = 5$ and $\sigma^2_B = 10$. {Circle}: Uniform noise with $u_G = 10$ and $u_B = 50$. {Triangle}: Speckle noise with $\sigma^2_G = 1$ and $\sigma_B^2 = 5\times10^{-2}$. {Rhombus}: Poisson noise with $\mu_G = 1$ and $\mu_B = 15$. We refer to equations \eqref{eq:gaussian_noise}, \eqref{eq:uniform_noise}, \eqref{eq:speckle_noise} and \eqref{eq:poisson_noise} for the definition of the different noise types.}
\label{tab:shape_parameters}
\end{table}

\begin{figure}
\centering
    \includegraphics[scale = 0.2]{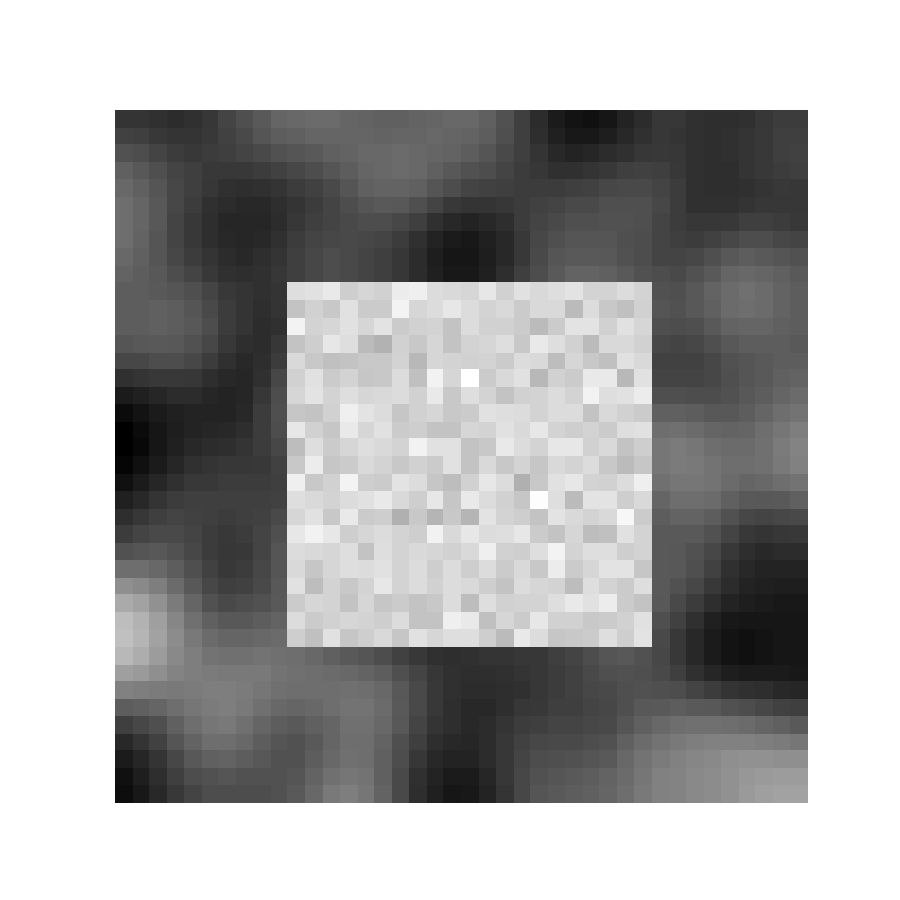}
    \includegraphics[scale = 0.2]{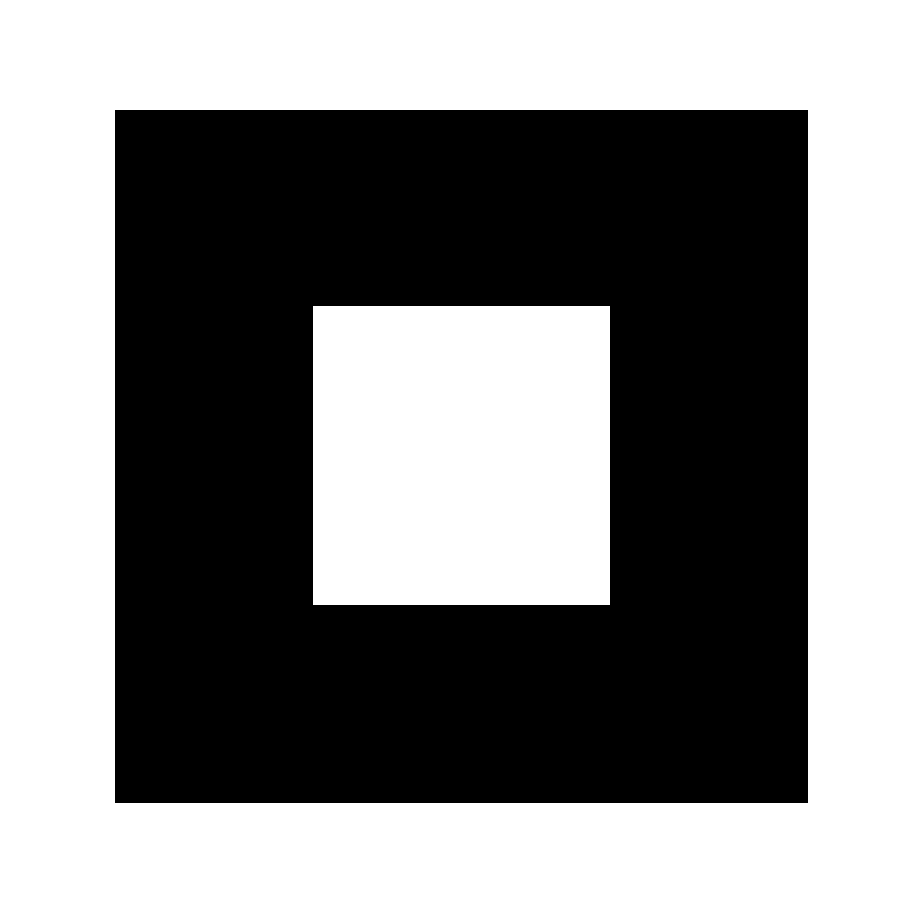}
    \includegraphics[scale = 0.2]{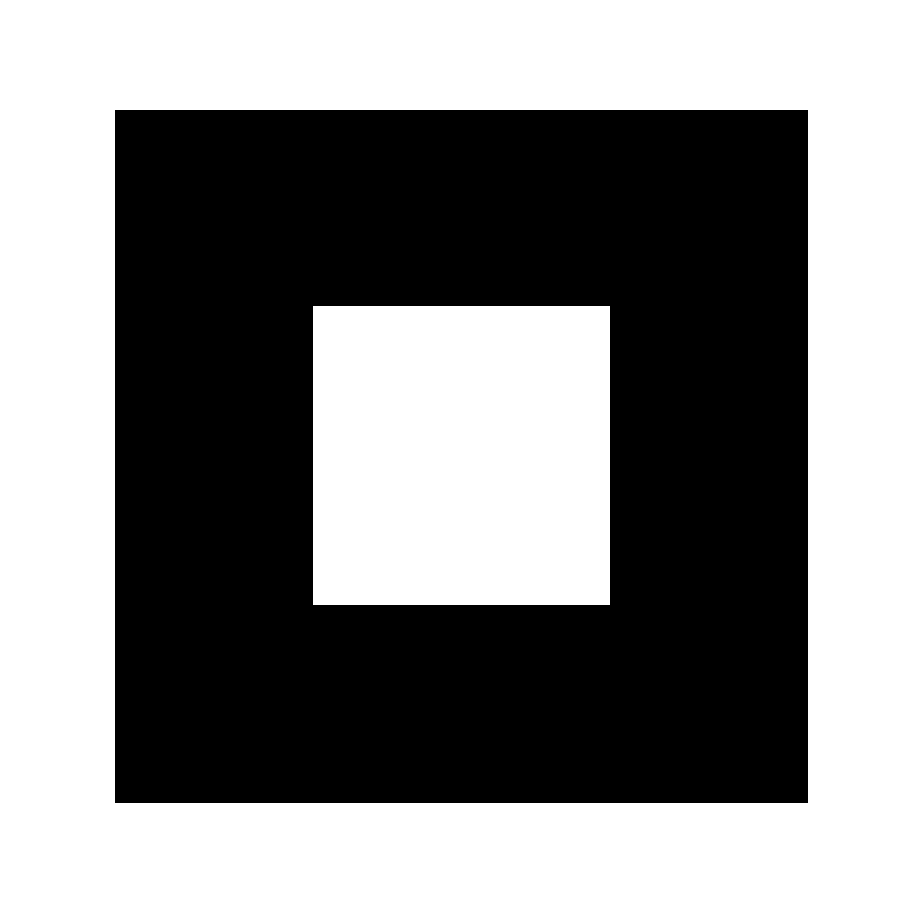} \\
    \includegraphics[scale  = 0.2]{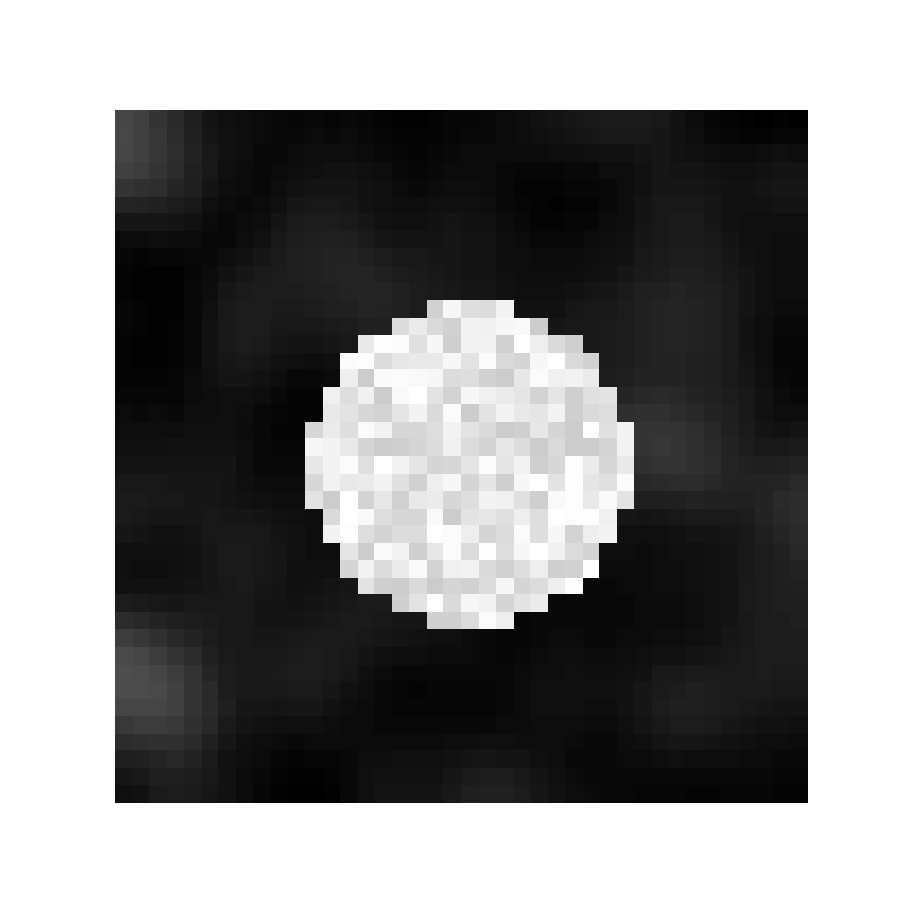}
    \includegraphics[scale = 0.2]{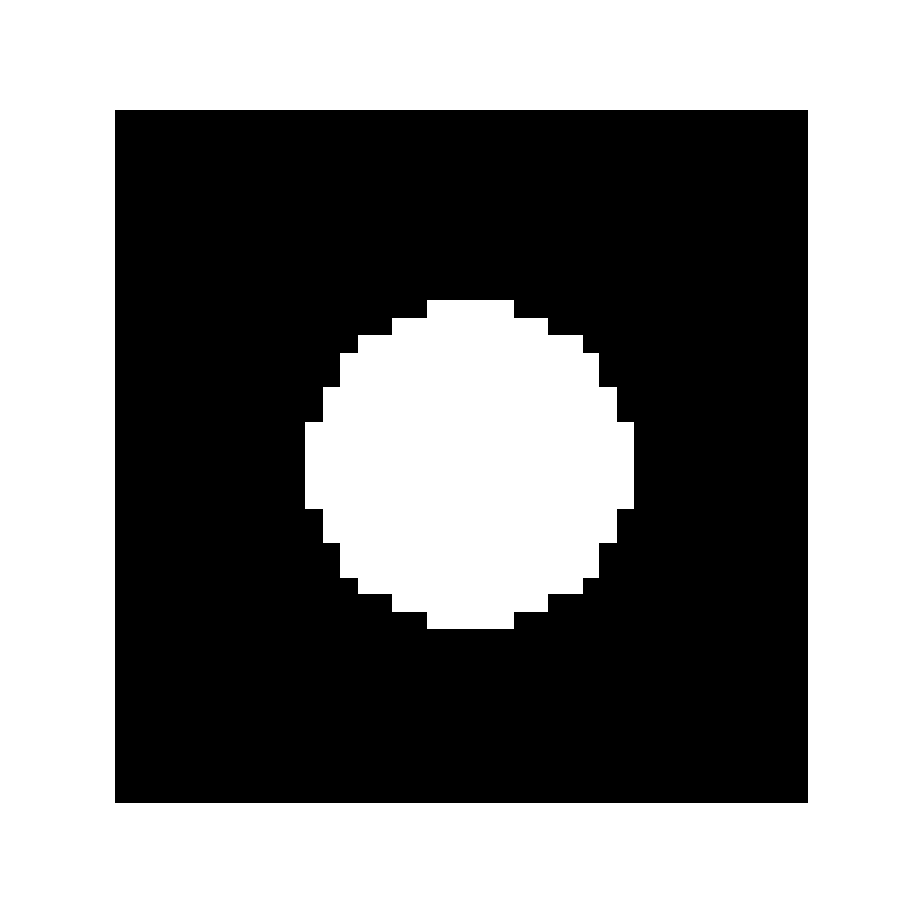}
    \includegraphics[scale = 0.2]{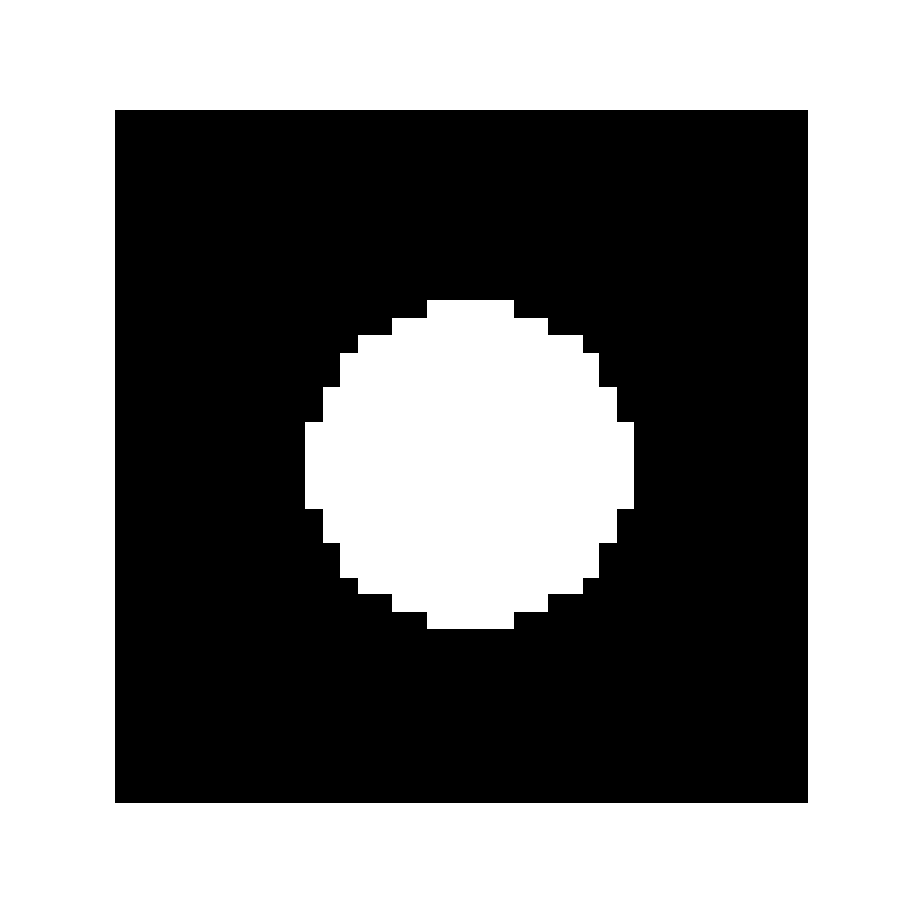}\\
    \includegraphics[scale = 0.2]{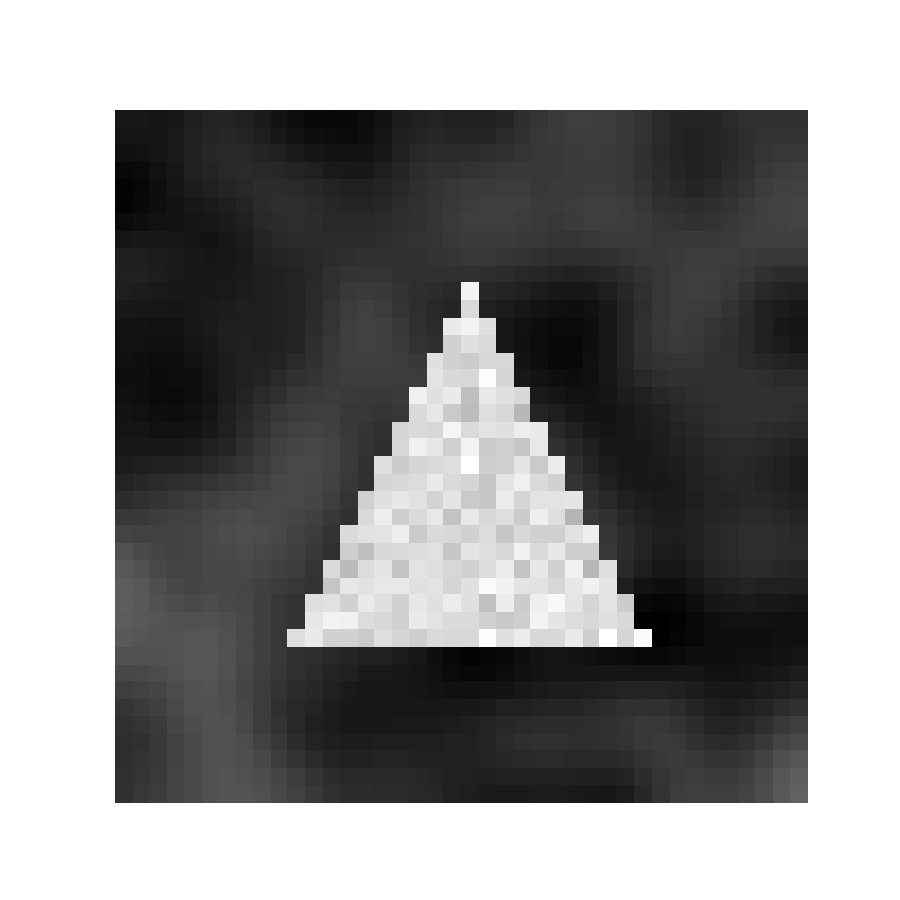}
    \includegraphics[scale = 0.2]{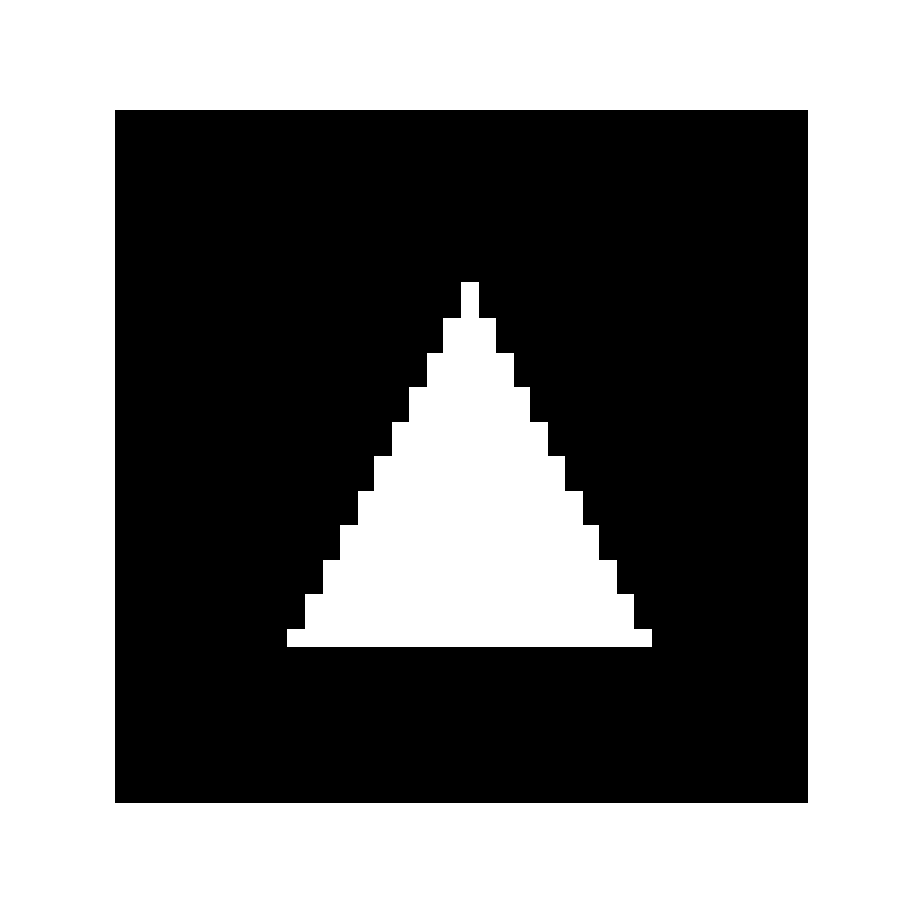}
    \includegraphics[scale = 0.2]{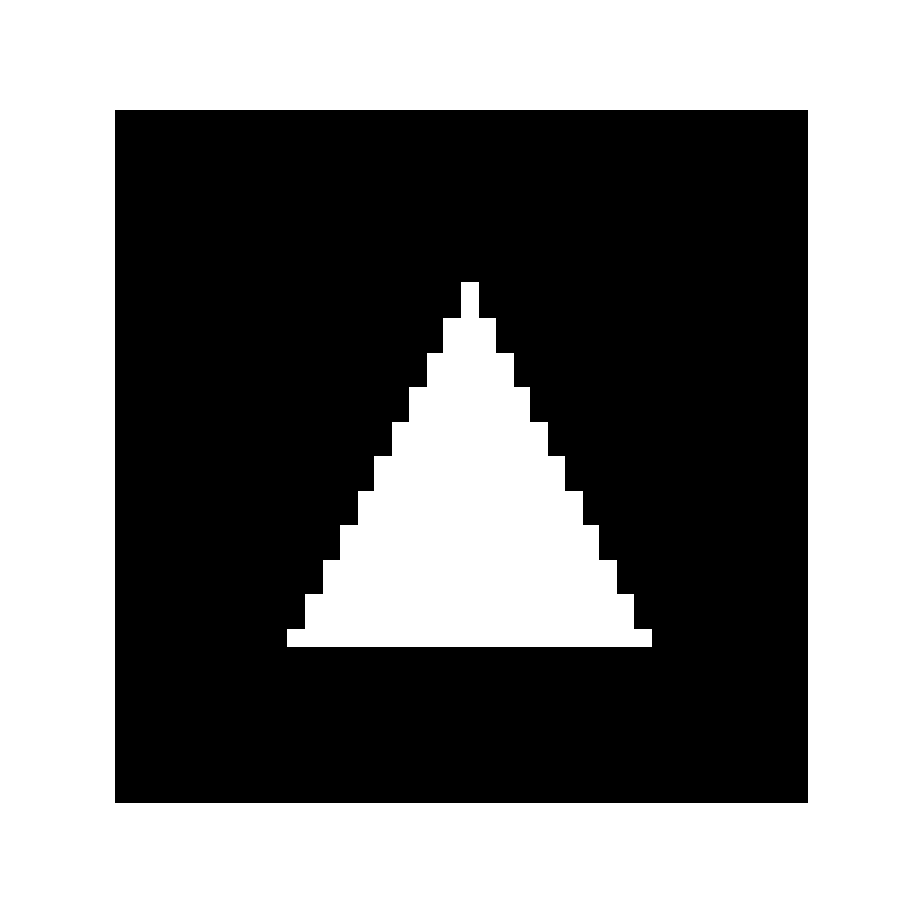}\\
    \includegraphics[scale = 0.2]{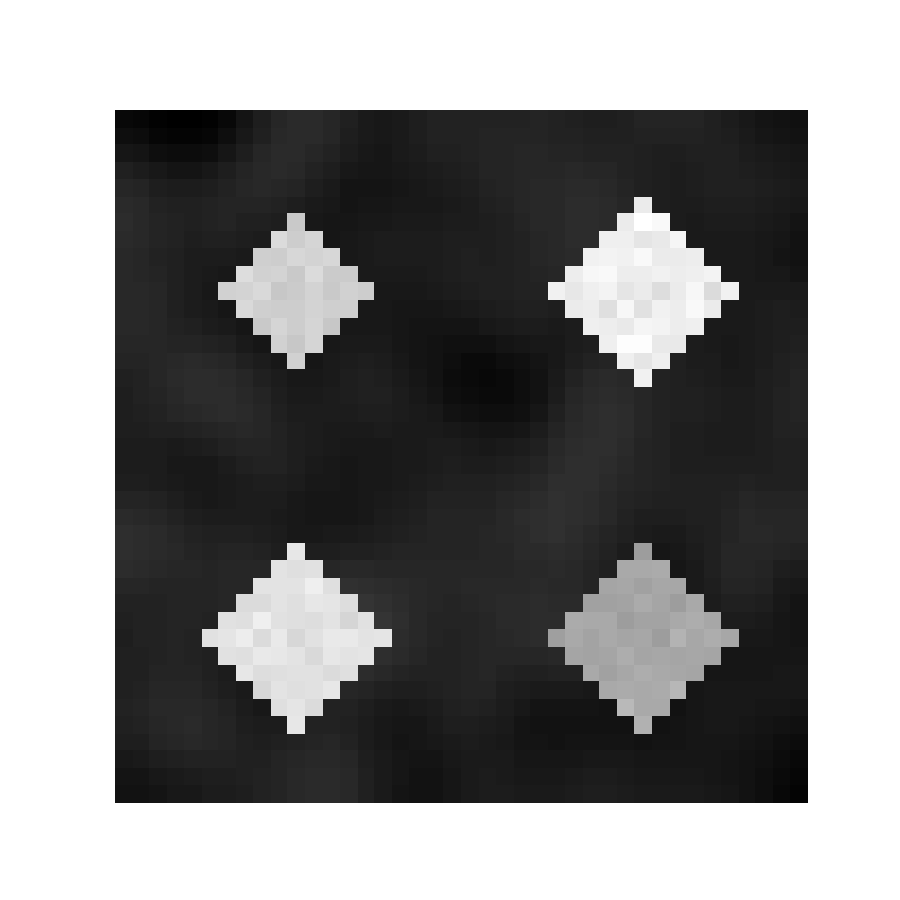}
    \includegraphics[scale = 0.2]{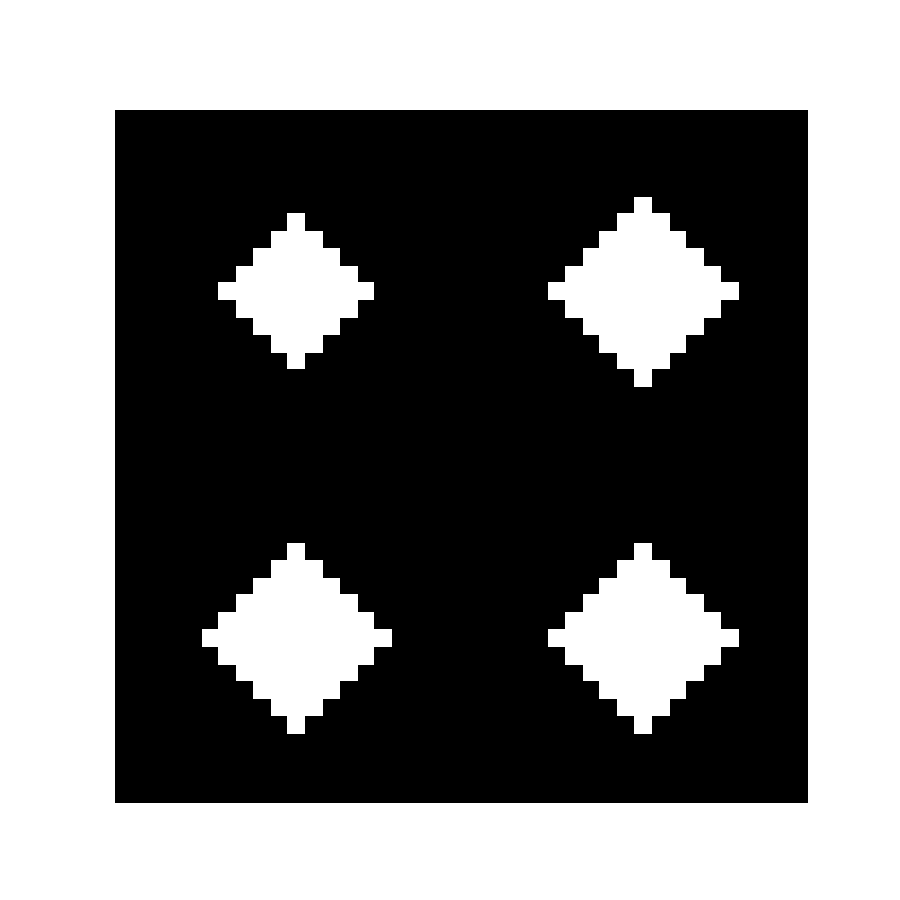}
    \includegraphics[scale = 0.2]{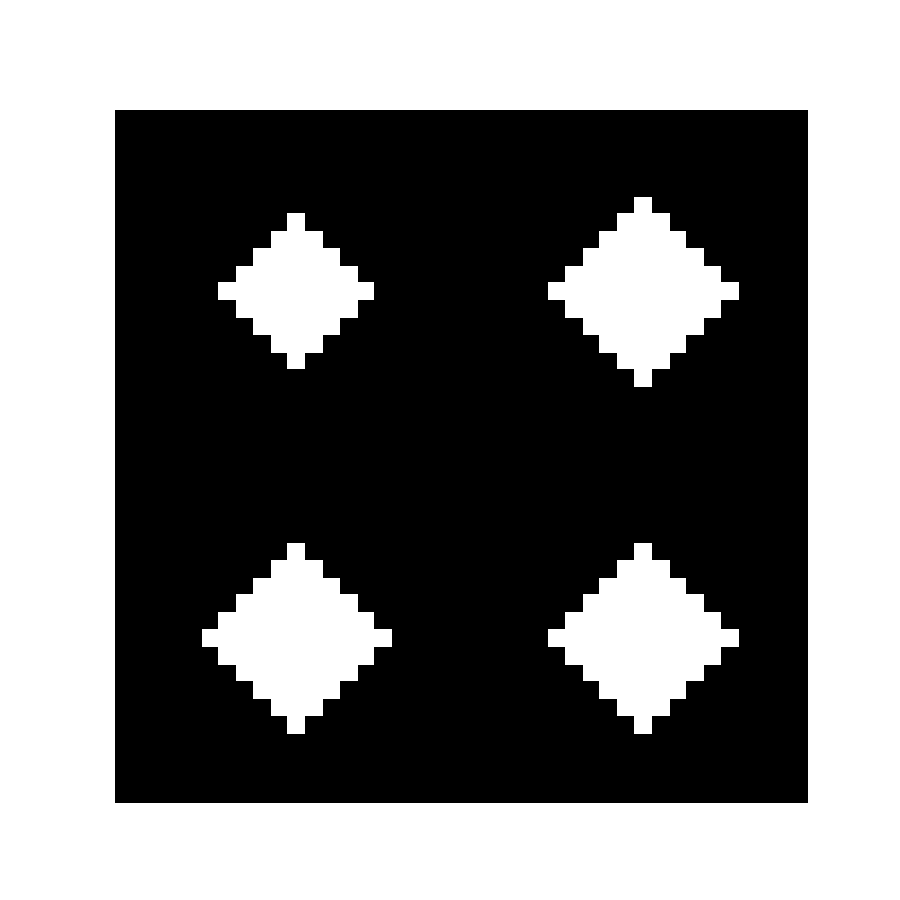}
\caption{\textbf{Test 3.1}.
{Top row:} Gaussian noise, centered square.
{Second row:} Uniform noise, centered circle.
{Third row:} Speckle noise, centered triangle.
{Fourth row:} Poisson noise, centered rhombus.
In each case, from left to right, we show the noisy image, the Ground Truth Segmentation Mask (GTSM), and the segmentation obtained with the optimized parameters.
}
\label{fig:segmentations}
\end{figure}

\begin{figure}
\centering
    \includegraphics[scale = 0.3]{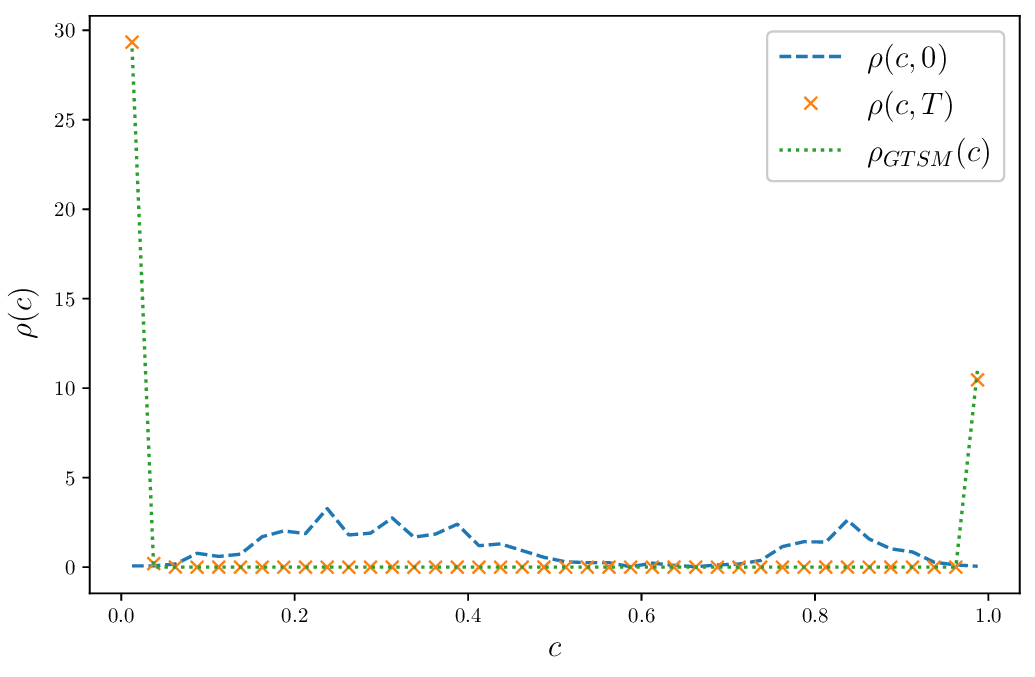}
    \includegraphics[scale = 0.3]{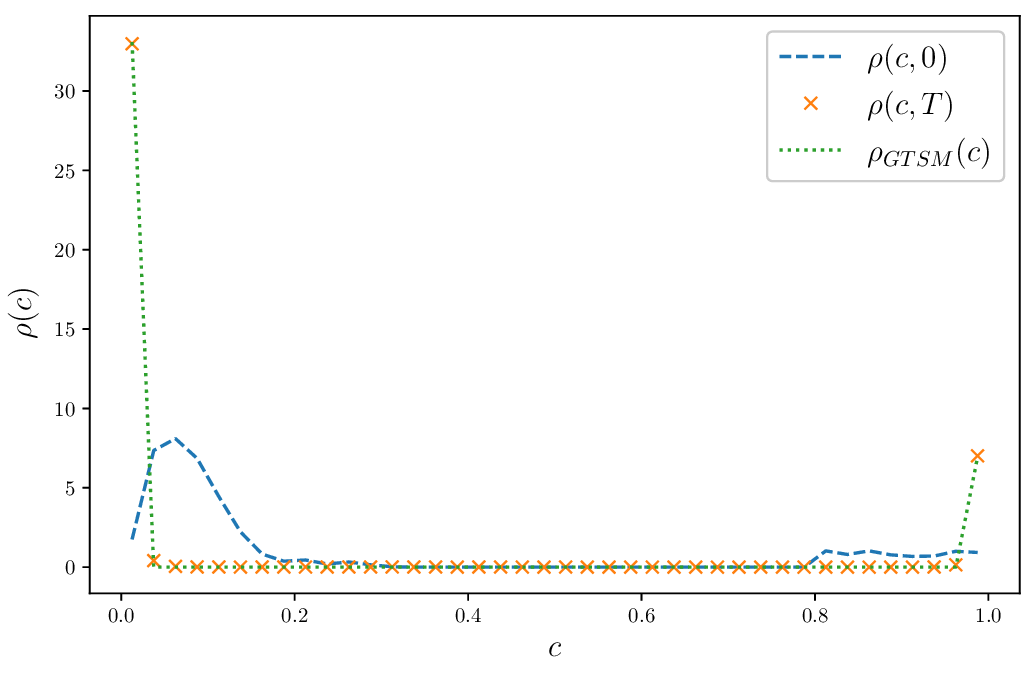}\\
    \includegraphics[scale = 0.3]{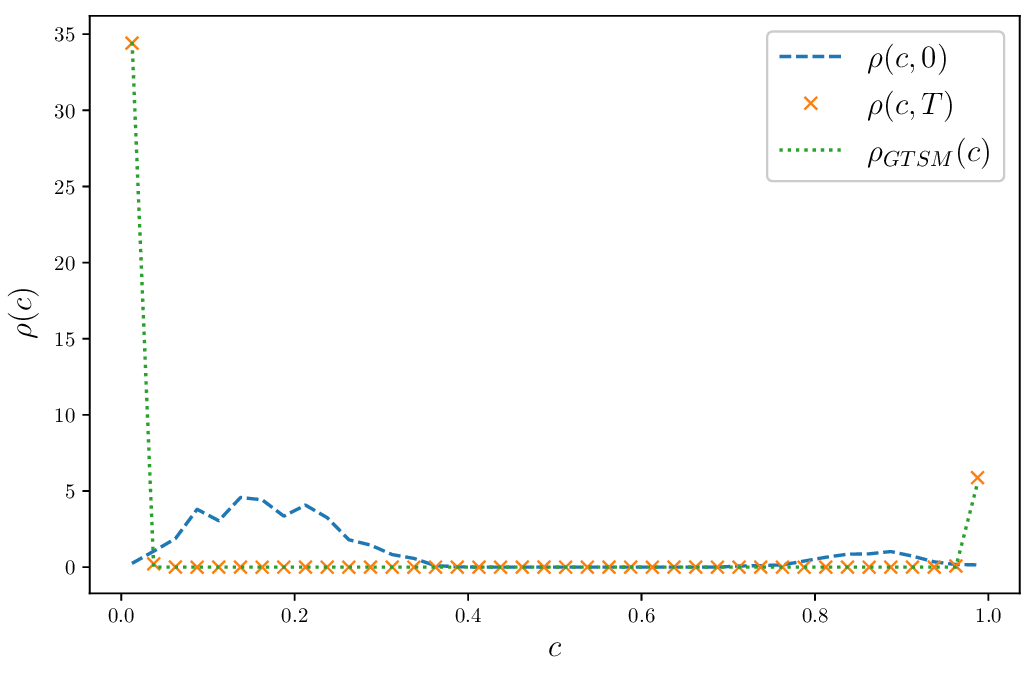}
    \includegraphics[scale = 0.3]{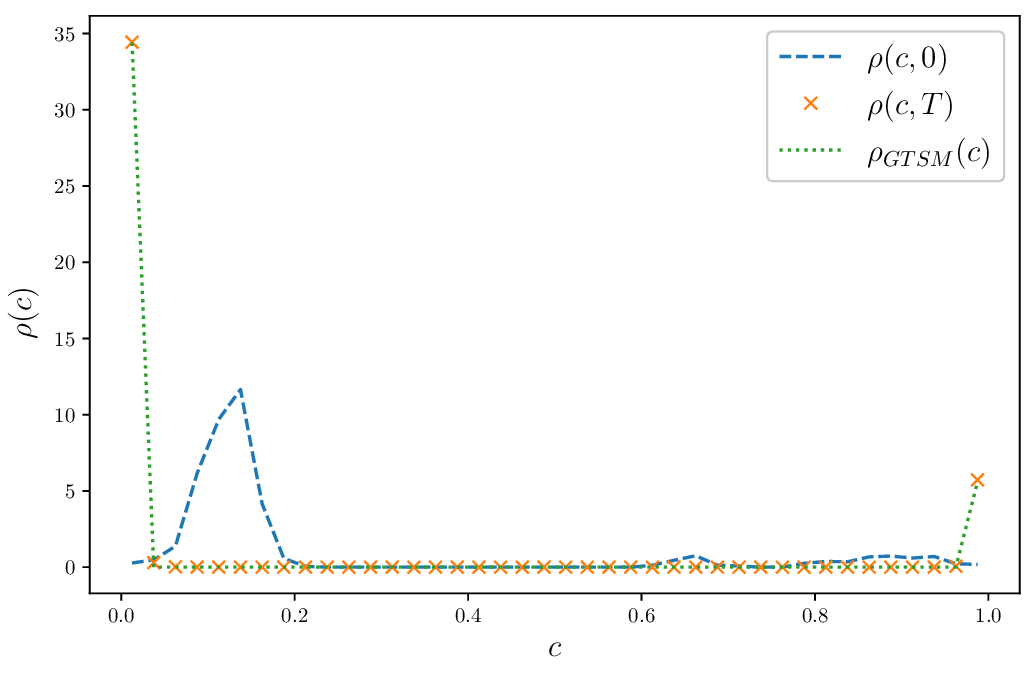}
\caption{\textbf{Test 3.1}. Feature density distributions for the different geometric shapes with varying noise configurations. In each case, we report the initial distribution $\rho(c,0)$, the Ground Truth Segmentation Mask (GTSM) $\rho_{\text{GTSM}}(c)$ and the large time distribution obtained from the macroscopic model $\rho(c,T)$ for $T=20$ obtained with the optimized parameters. {Top Left:} Gaussian noise, centered square. {Top Right:} Uniform noise, centered circle. {Bottom Left:} Speckle noise, centered triangle. {Bottom Right:} Poisson noise, rhombus.}
\label{fig:densities}
\end{figure}

The original images, the ground truth segmentation mask and the segmentations obtained are reported in Figure~\ref{fig:segmentations}. The optimal parameters, obtained as discussed in Section \ref{sect:test3}, are reported in Table \ref{tab:shape_parameters}. In it we can observed that the set of optimal parameters obtained are able to correctly reproduce the expected segmentation mask.
Furthermore, Figure~\ref{fig:densities} illustrates the different image profiles at the level of the feature density distributions. We can observed the initial distribution of the feature $\rho(c,0)$, the GTSM segmentation mask $\rho_{\text{GTSM}}(c)$ and the large time distribution obtained from the macroscopic model  $\rho(c,T)$ with the optimized parameters for $T=20$ with $N_c = 30$ feature points. We notice how the set of optimal parameters correctly distributes the feature density to match the expected distribution mask with robust performance with respect to the different shapes. Finally, we report the optimal parameters and the loss obtained for each of the geometric shapes in Table~\ref{tab:shape_parameters}.

\subsubsection{Test 3.2: impact of different noise distributions and levels}

In the following set of tests, we focus on the segmentation of the central square under different noise types and intensities. In these experiments, the same noise intensity is applied to both the foreground and background regions. In this case we set as final time $T=50$ and consider $N_c = 30$ points in the feature space.

\begin{figure}
\centering
\includegraphics[scale = 0.3]{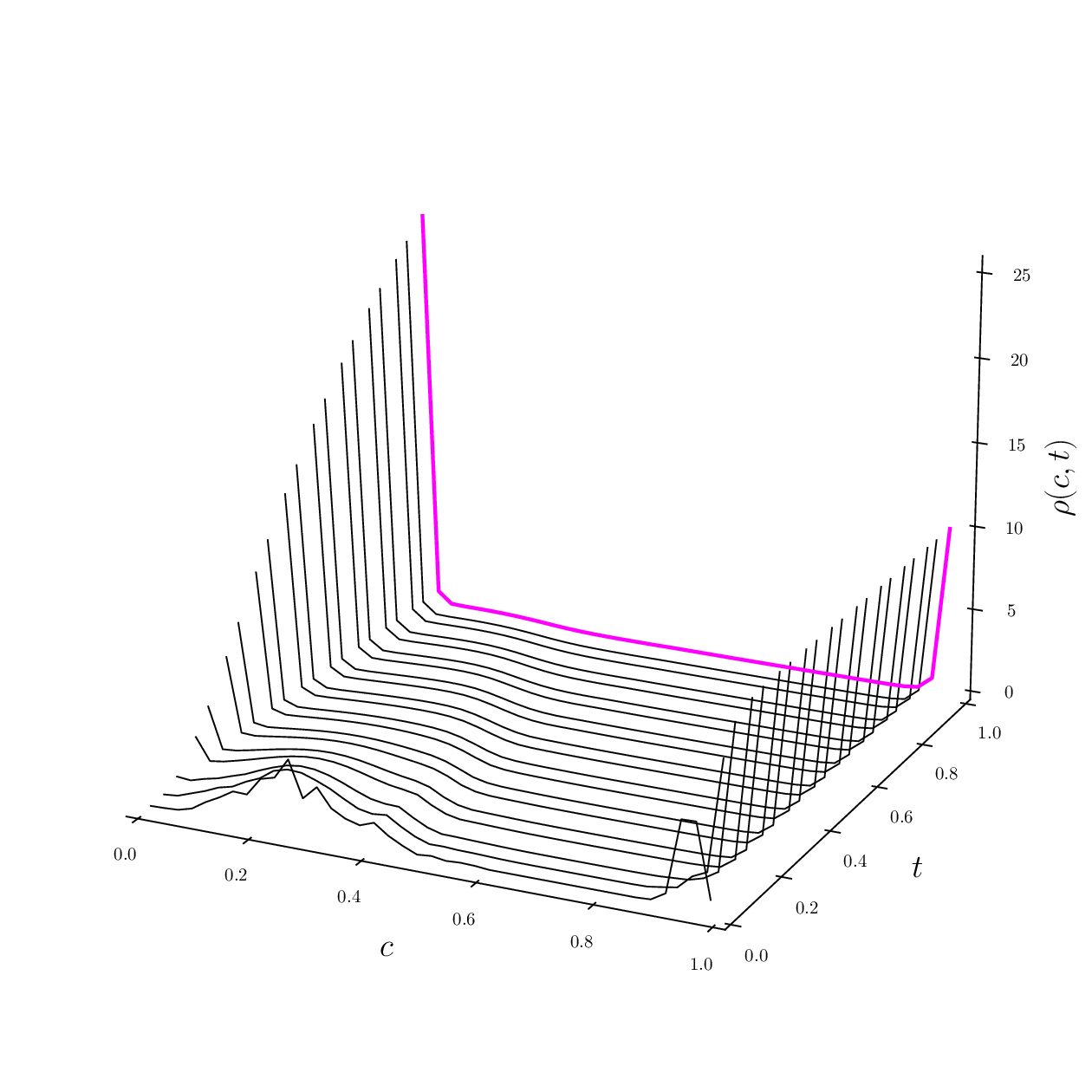}
\includegraphics[scale = 0.3]{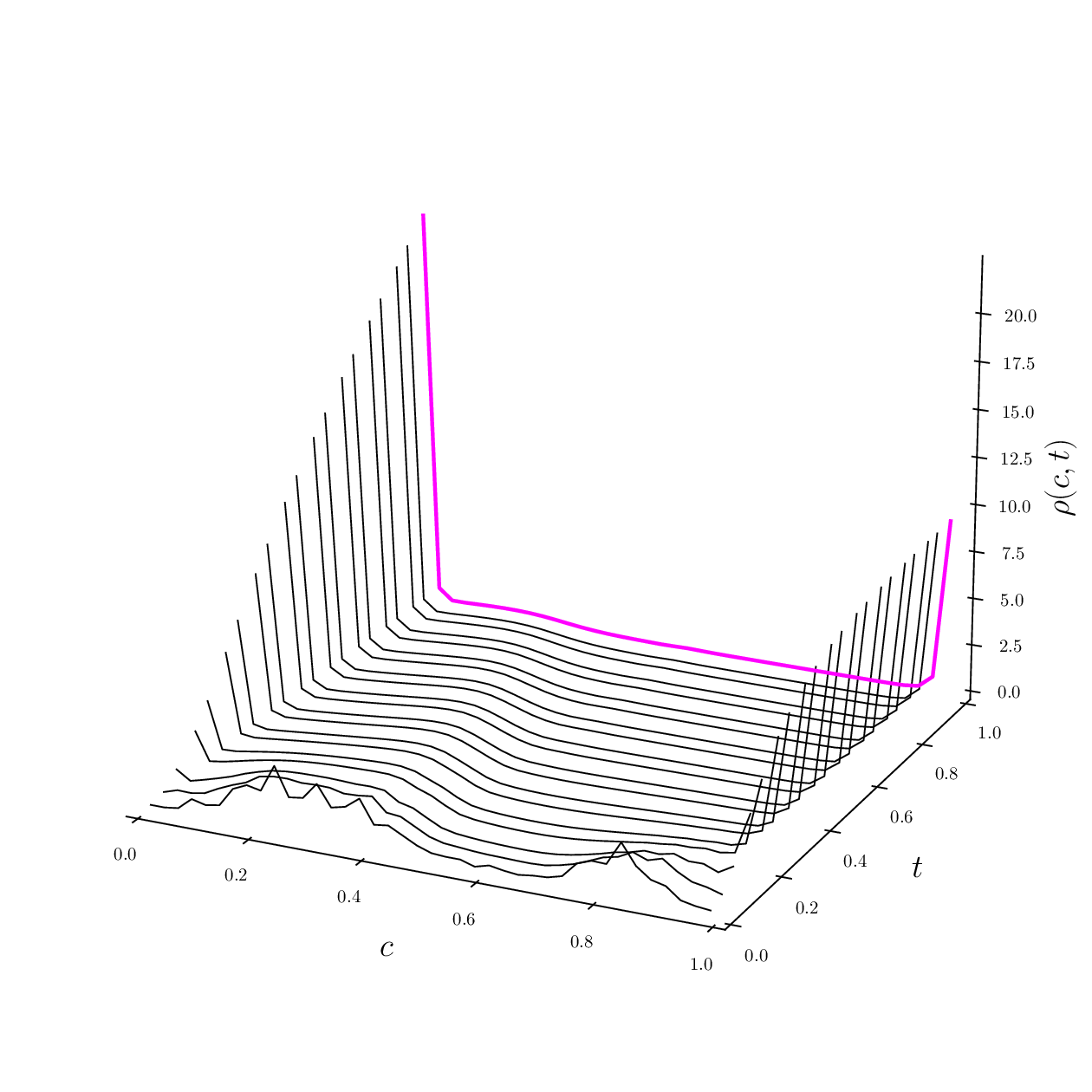}\\
\includegraphics[scale = 0.3]{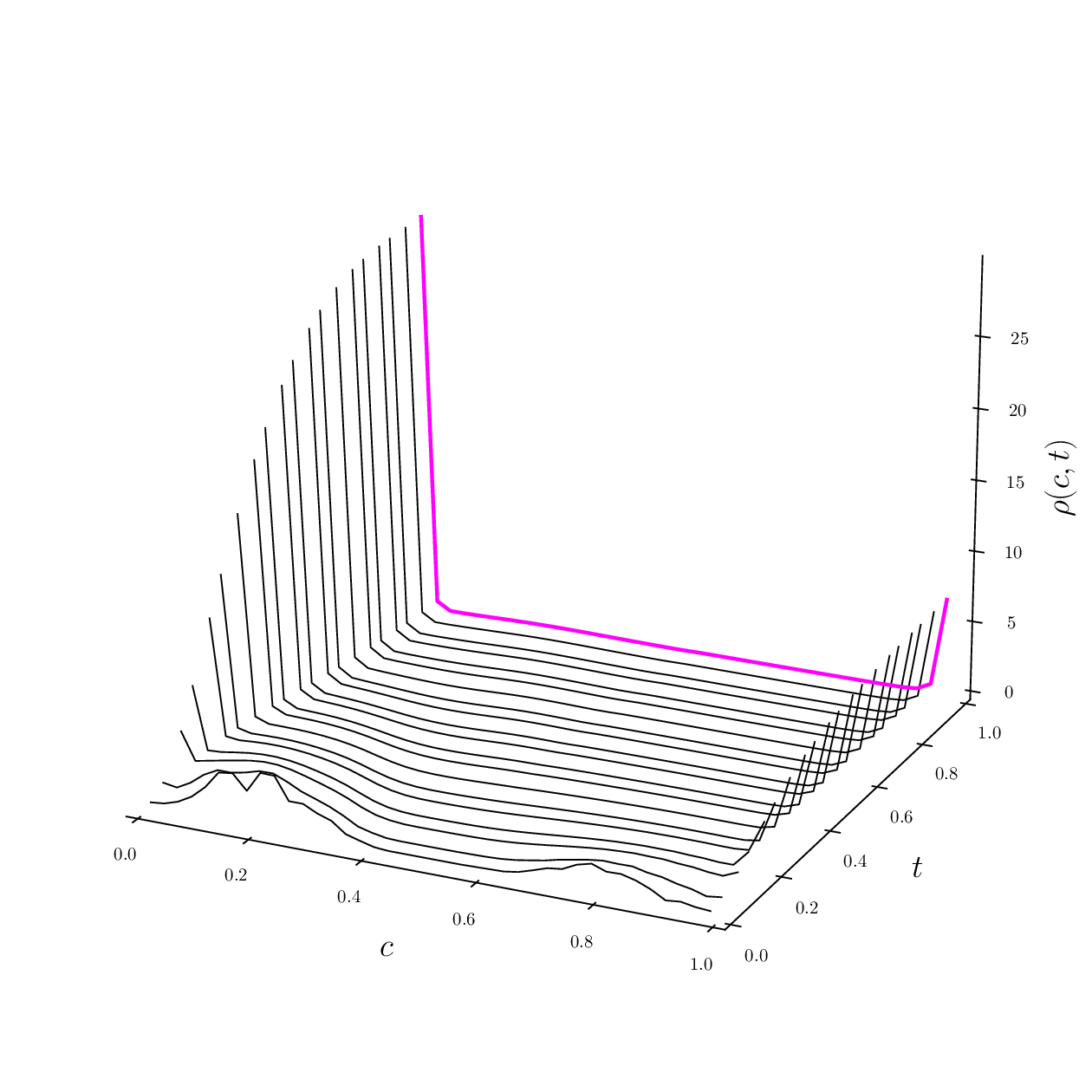}
\includegraphics[scale = 0.3]{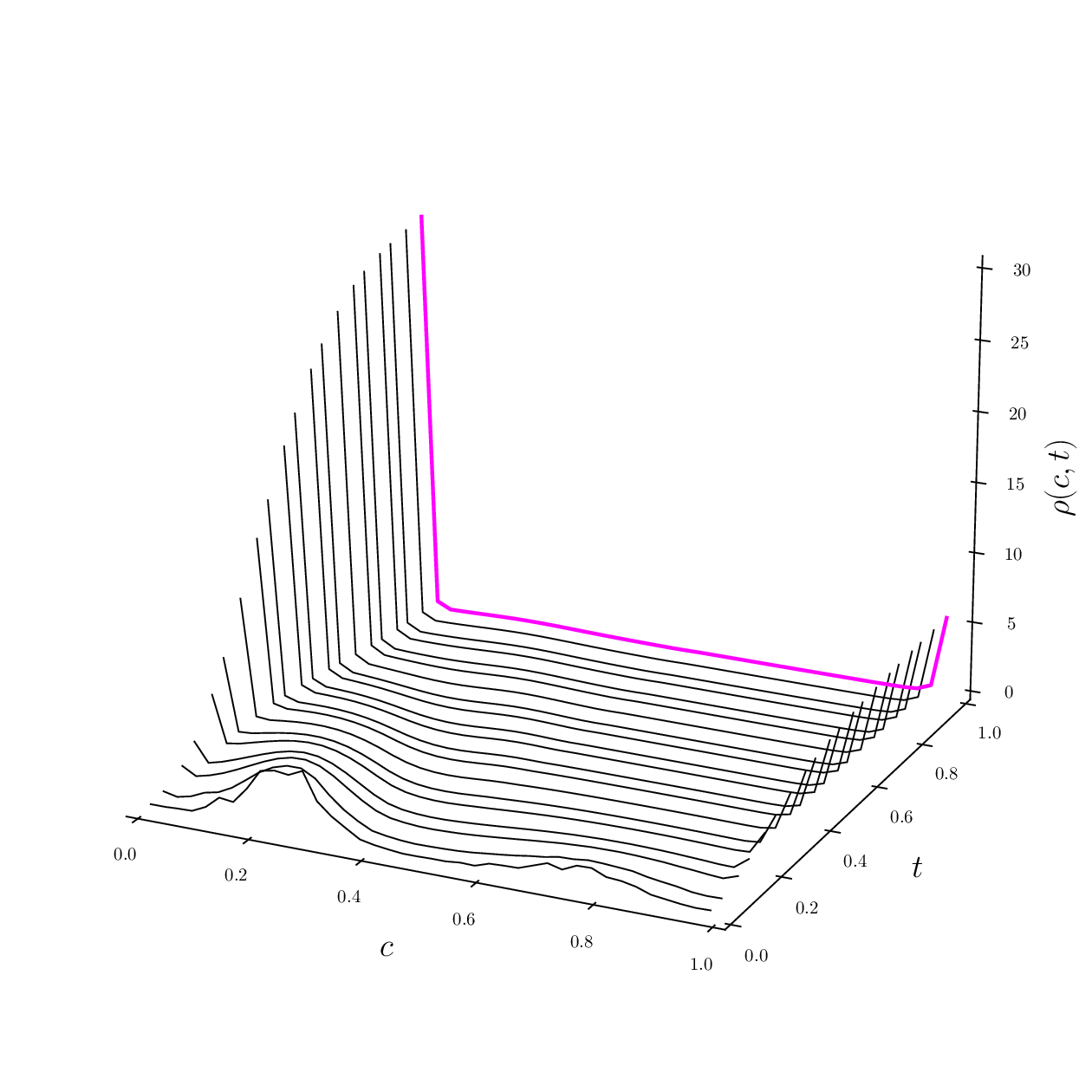}
\caption{\textbf{Test 3.2}. We report the evolution of the density $\rho(c,t)$ for four square images with progressively increasing Gaussian noise levels ($\sigma^2 = 2,5,10,15$). The optimized parameters effectively capture the expected behaviour of the GTSM, redistributing the density such that it concentrates near $c=0,1$.}
\label{fig:rho_evolution}

\end{figure}

Table~\ref{tab:square_parameters} reports the loss values obtained for the various noise configurations. In all cases, the resulting segmentation mask correctly identifies the target region, although differences in the loss values can be observed at the distribution level. The resulting segmented image are not shown to avoid repetition. Furthermore, in Figure~\ref{fig:rho_evolution} we report the evolution of the density $\rho(c,t)$ for the case of the Gaussian noise with $\sigma^2_G = 2,5,10,15$ with the optimized parameters for short times. It can be seen how the different noise intensities affect the inital condition of the density which in short time the distribution binarizes concentrating all the respective mass towards $c=0,1$. 

\begin{table}
\centering
\begin{tabular}{ccccc}
\hline
\multicolumn{1}{|l}{}&\multicolumn{4}{|c|}{Loss}\\
\hline
\multicolumn{1}{|l|}{Intensity} & \multicolumn{1}{l|}{Gaussian} & \multicolumn{1}{l|}{Poisson} & \multicolumn{1}{l|}{Uniform} & \multicolumn{1}{l|}{Speckle} \\ \hline
\multicolumn{1}{|c|}{1}         & \multicolumn{1}{c|}{0.0135}   & \multicolumn{1}{c|}{0.0781}  & \multicolumn{1}{c|}{0.0781}  & \multicolumn{1}{c|}{0.0066}  \\ \hline
\multicolumn{1}{|c|}{2}         & \multicolumn{1}{c|}{0.0226}   & \multicolumn{1}{c|}{0.0755}  & \multicolumn{1}{c|}{0.0780}  & \multicolumn{1}{c|}{0.0061}  \\ \hline
\multicolumn{1}{|c|}{3}         & \multicolumn{1}{c|}{0.1209}   & \multicolumn{1}{c|}{0.0887}  & \multicolumn{1}{c|}{0.0766}  & \multicolumn{1}{c|}{0.0636}  \\ \hline
\multicolumn{1}{|c|}{4}         & \multicolumn{1}{c|}{0.1595}   & \multicolumn{1}{c|}{0.0888}  & \multicolumn{1}{c|}{0.0752}  & \multicolumn{1}{c|}{0.0876}  \\ \hline
\multicolumn{1}{l}{}            & \multicolumn{1}{l}{}          & \multicolumn{1}{l}{}         & \multicolumn{1}{l}{}         & \multicolumn{1}{l}{}        
\end{tabular}
\caption{\textbf{Test 3.2}. We report the loss values obtained from \eqref{eq:opt} for the segmentation of the square image with different noise types and intensities. Gaussian: $\sigma^2 = 2,5,10,15$ ; Poisson: $\mu = 0.01,0.05,0.1,0.2$ ; Uniform: $u = 2,5,10,15$ ; Speckle: $\sigma^2 = 0.01,0.02,0.03,0.04$. We point out that the same intensity is applied to both the foreground and background regions. We refer to equations \eqref{eq:gaussian_noise}, \eqref{eq:uniform_noise}, \eqref{eq:speckle_noise} and \eqref{eq:poisson_noise} for the definition of the different noise types.}
\label{tab:square_parameters}
\end{table}

\begin{figure}
\centering
\includegraphics[width=\textwidth]{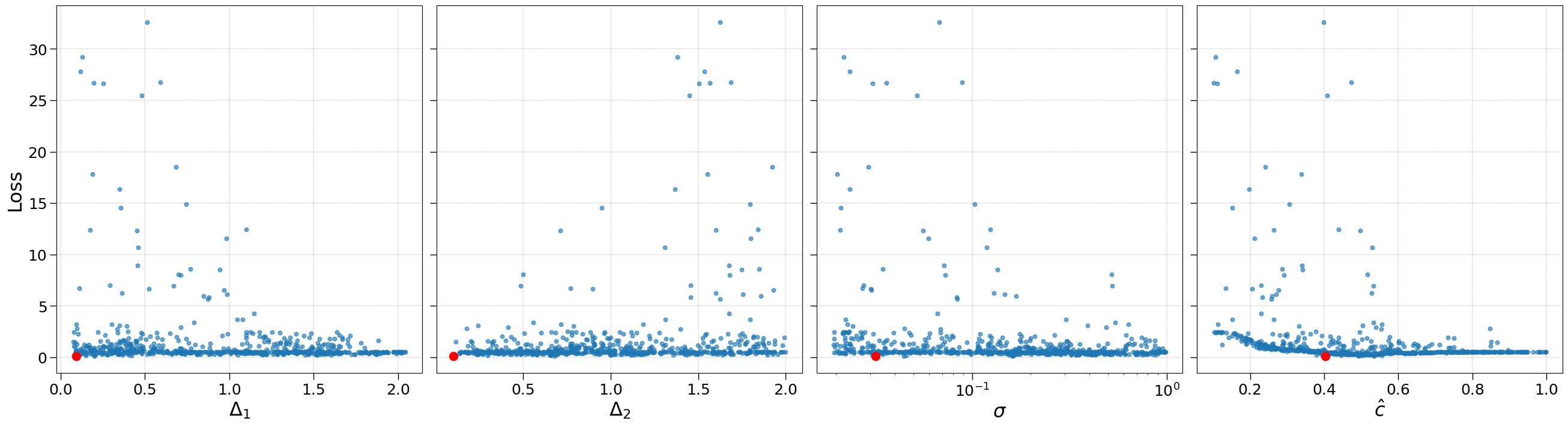}
\caption{\textbf{Test 3.2}. We report the loss values obtained for the different combination of the model parameters for the case of the Speckle noise with $\sigma^2 = 0.04$. We notice the difficulty this optimization problem poses given that the loss landscape is non-convex and flat on large regions of the parameter space which makes it very likely to get stuck in local minima. The red point represents the optimal combination of parameters obtained with the CBO optimization algorithm.}
\label{fig:cbo_tuning}
\end{figure}

Lastly, in Figure~\ref{fig:cbo_tuning} we depict the loss landscape corresponding to the Speckle noise case with $\sigma^2 = 0.04$, as obtained through the optimization procedure. The figure highlights the non-convex nature of the optimization problem and illustrates the effectiveness of the Consensus-Based Optimization (CBO) approach in identifying suitable parameters. Despite the non-convexity of the loss landscape, the optimization procedure consistently identified parameter configurations leading to the correct segmentation mask across all considered noise scenarios.

\subsection{Test 4: RGB case}

In this test, we extend the previous setting for the segmentation of color images. We now consider a three-dimensional feature space, where each component of the feature vector corresponds to a channel of the RGB color space. In this setting, $\rho(c):\mathbb{R}^3 \to \mathbb{R}$ represents the feature density in the three-dimensional color space and we define the marginal distributions as

\begin{equation}
    \rho_i(c_i) = \int_{\mathbb{R}^2} \rho(c_r,c_g,c_b) dc_j dc_k, \qquad i,j,k \in \{R,G,B\}, i\neq j\neq k.
\end{equation}

\begin{figure}
\centering
\includegraphics[width = 4cm]{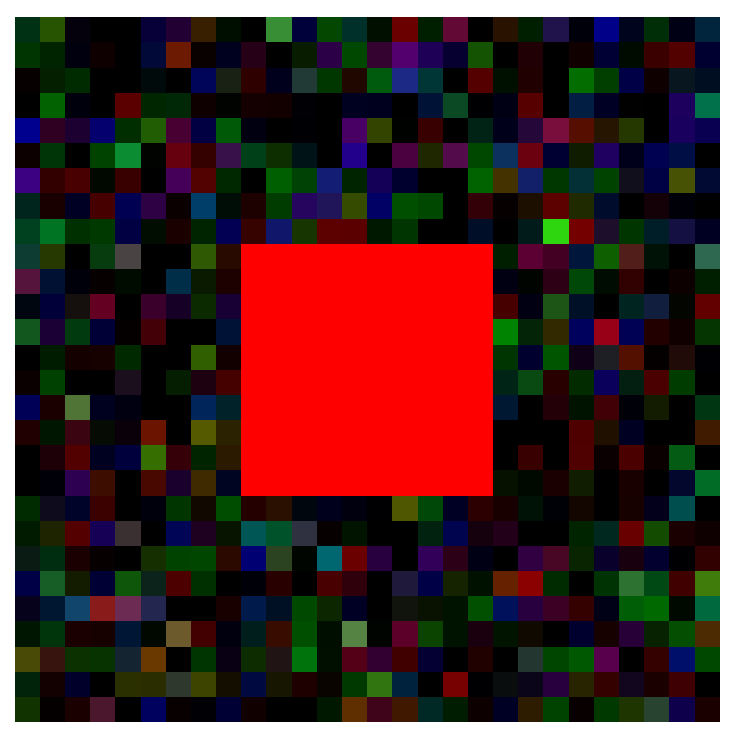}
\includegraphics[width = 4cm]{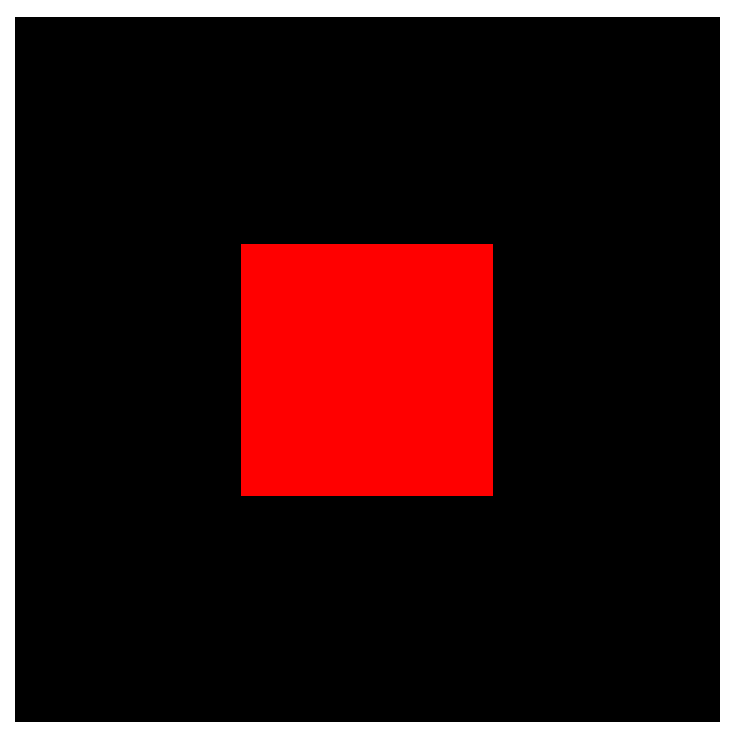}
\includegraphics[width = 4cm]{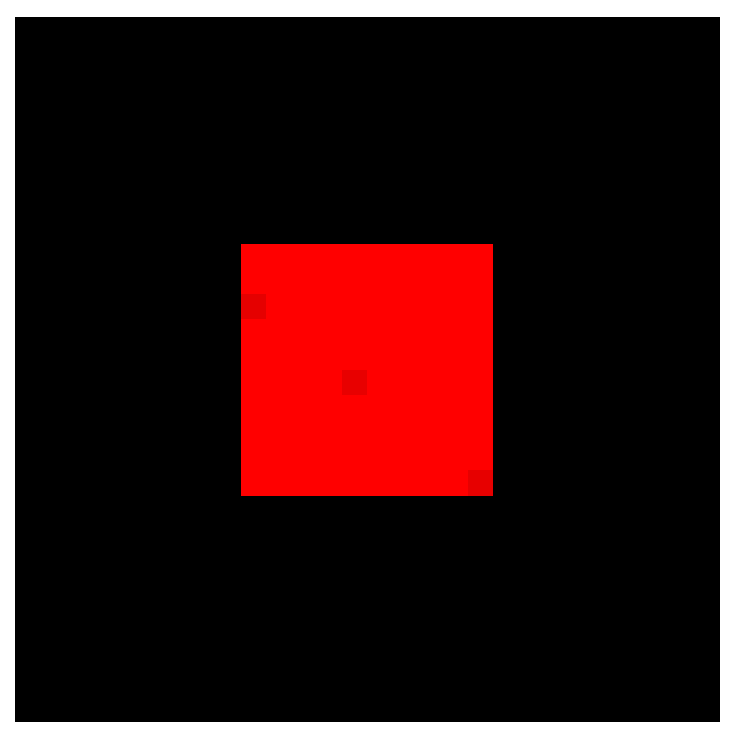}\\
\includegraphics[width = 4cm]{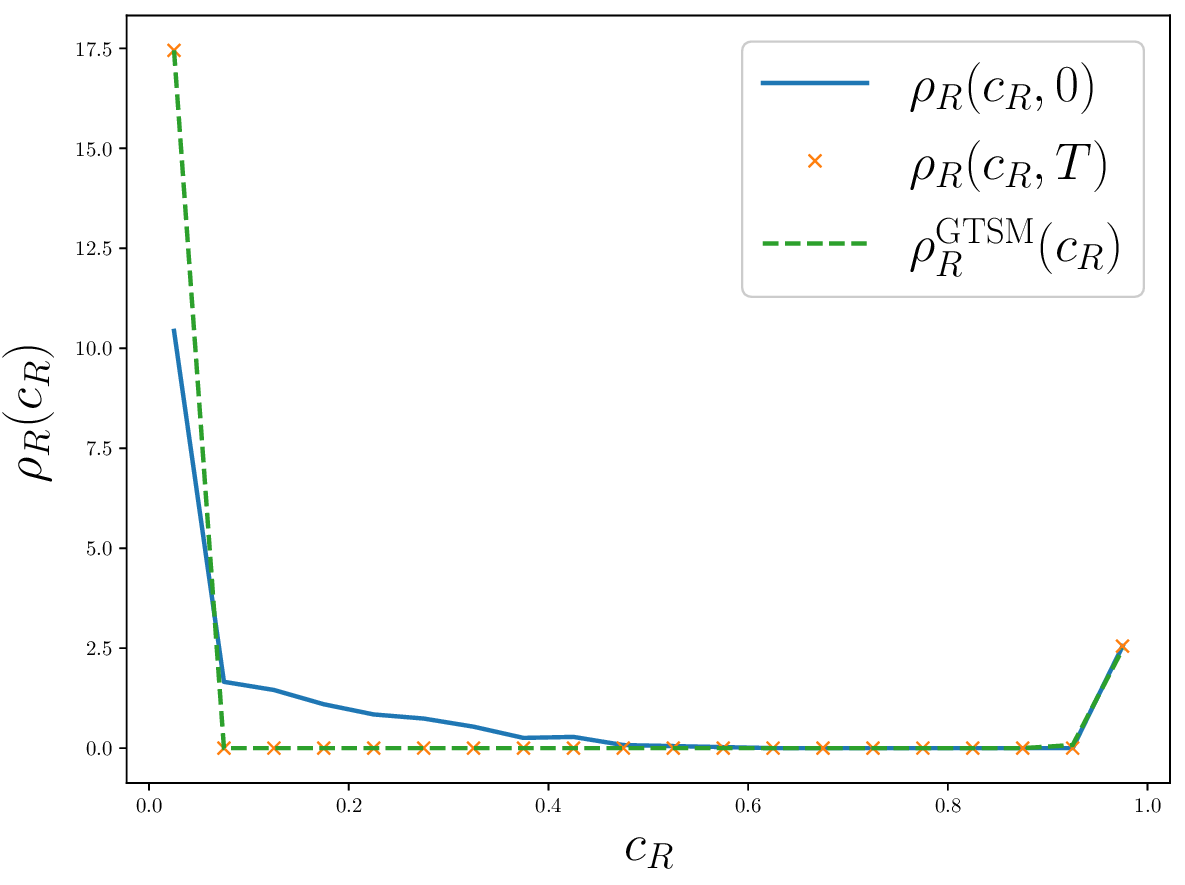}
\includegraphics[width = 4cm]{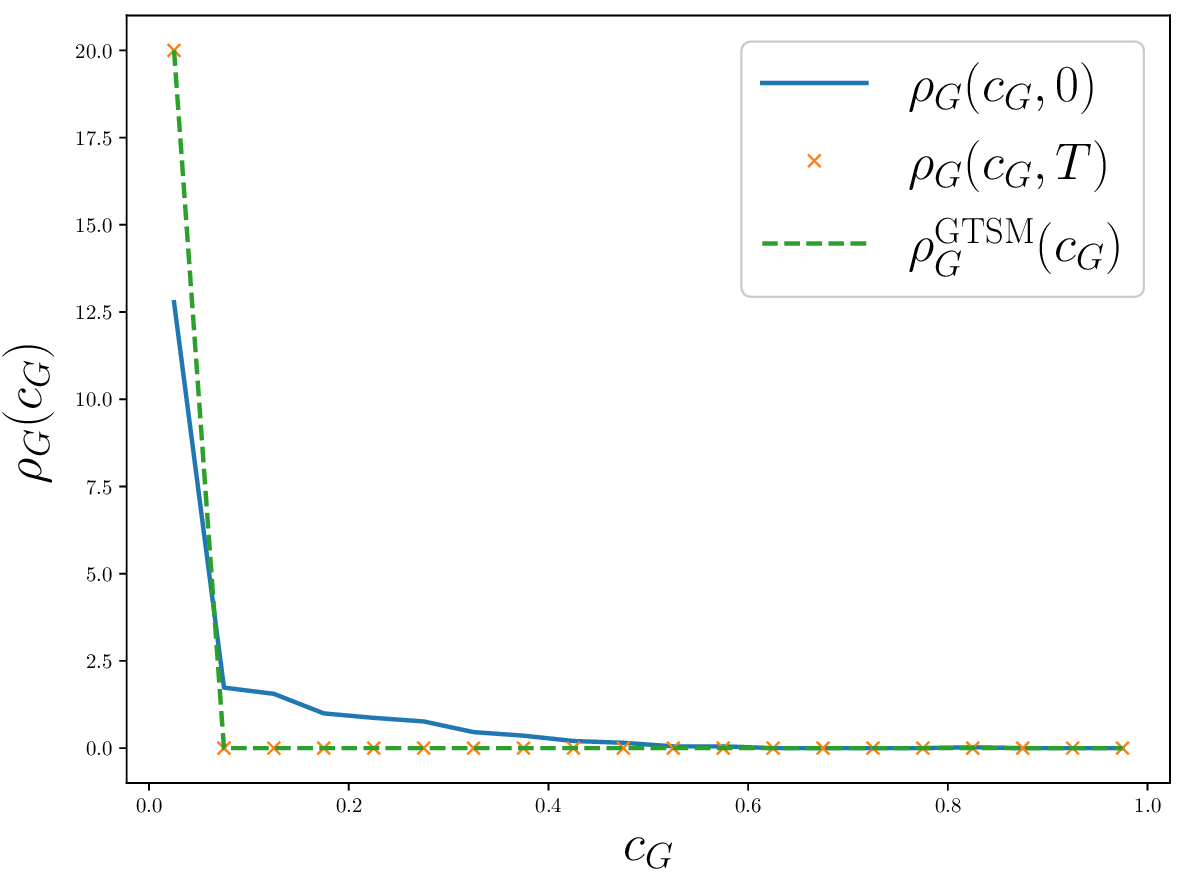}
\includegraphics[width = 4cm]{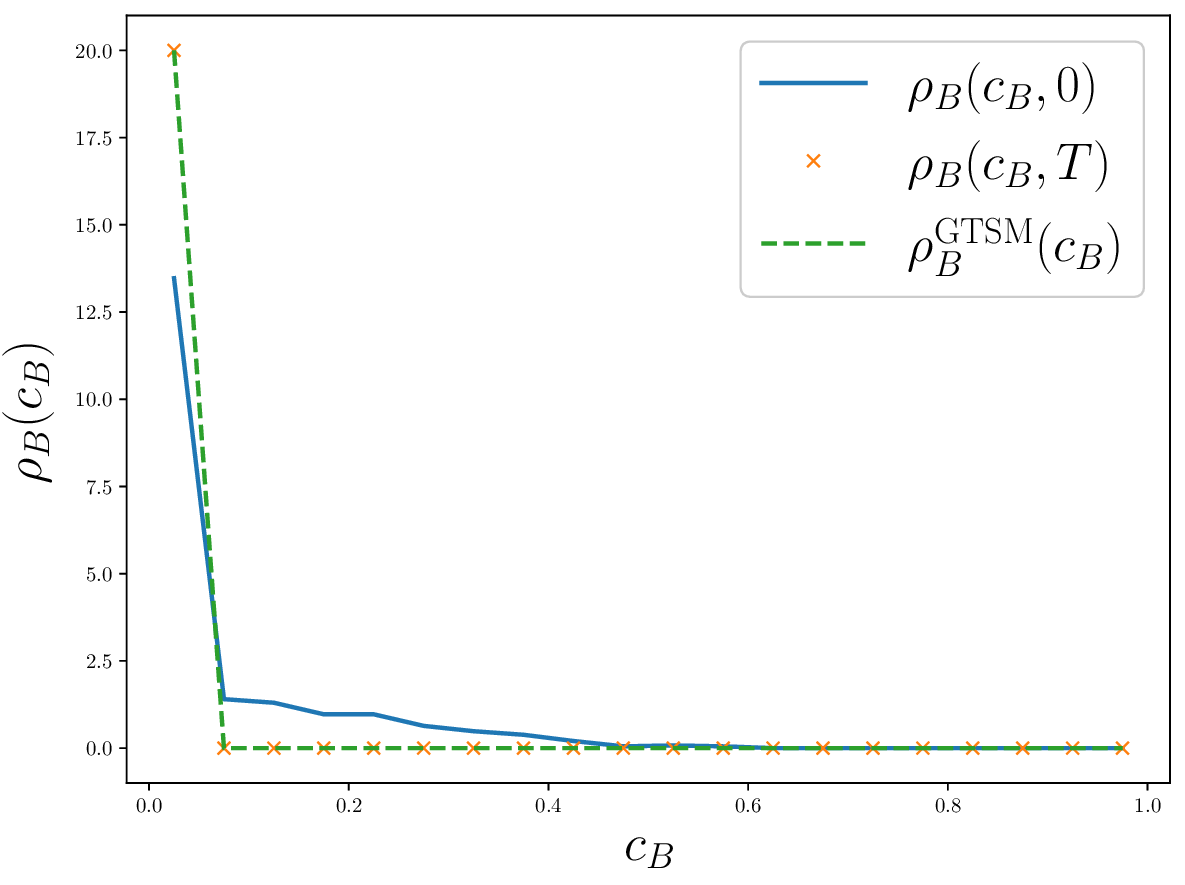}
\caption{\textbf{Test 4}. Segmentation of a $28\times 28$ color image with Gaussian noise in all channels with intensity $\sigma^2_R = \sigma^2_G = \sigma^2_B = 0.2$. {Top}: We show the original image, the Ground Truth Segmentation Mask (GTSM), and the segmentation obtained with the optimal parameters. {Bottom}: We report the marginal distributions $\rho_R(c_R)$, $\rho_G(c_G)$, and $\rho_B(c_B)$ obtained with the optimal parameters compared to the marginals of $\rho_{\text{GTSM}}$. We observe that the optimization procedure was able to find a set of parameters such that the asymptotic distribution converges to the expected distribution in all three channels.}
\label{fig:color_image}

\end{figure}

In the following we consider the following noise structure for a color image as 

\begin{equation}
\tilde{I}(x,y,c) = \begin{cases}
\mathcal{G}^c_G(I),\quad \text{if } (x,y) \in I_G \\
\mathcal{G}^c_B(I),\quad \text{if } (x,y) \in I_B
\end{cases}
\end{equation}

where $I_G$ and $I_B$ represent the shape and the background regions, respectively, and $\mathcal{G}^c_G(I)$ and $\mathcal{G}^c_B(I)$ represent the noise structure applied to the shape and the background respectively for each channel $c \in \{R,G,B\}$. In this way, again, we can consider different noise types and intensities for each channel and for the shape and the background. We consider a $28\times 28$ color image with a centered square and a gaussian background noise which we applied to all channels with intensity $\sigma^2_R = \sigma^2_G = \sigma^2_B = 0.2$ and final time $T = 400$ with $N_c=30$ for all the three color directions. As optimization procedure we consider the same CBO method described in the previous section with the same settings. Given that our image is composed of $784$ pixels, to reduce the stochasticity associated with the particles dynamics used to generate the segmentation mask, we consider the segmentation obtained to be the average mask after $10$ independent runs of the Algorithm~\ref{alg:DSMC_complete}. The results obtained are shown in Figure~\ref{fig:color_image}. We observe that the resulting segmentation mask is accurately reconstructed. Furthermore, we present the initial, asymptotic, and target marginal distributions for each channel. We observe that the optimization procedure successfully identifies a set of parameters for which the asymptotic distribution matches the target distribution across all three channels. 
For this experiment, the optimal parameters and the corresponding loss value are reported in Table~\ref{tab:optimal_parameters_color}.

\begin{table}
\centering
\begin{tabular}{|c|c|c|c|c|c|c|}
\hline
$\Delta_1$
& $\Delta_2$
& $\sigma^2$
& $\tilde{c}_R$
& $\tilde{c}_G$
& $\tilde{c}_B$
& Loss \\
\hline
0.5742
& 0.5016
& 0.0493
& 0.5413
& 0.5488
& 0.5533
& $0.0273$ \\
\hline
\end{tabular}
\caption{\textbf{Test 4}. Optimal parameters for the color image with Gaussian noise in all channels with intensity $\sigma^2_R = \sigma^2_G = \sigma^2_B = 0.2$.}
\label{tab:optimal_parameters_color}
\end{table}

\section{Conclusion}\label{sect:conclusions}

In this work, we introduced new a multiscale kinetic framework for consensus-based image segmentation. Starting from a decoupled interaction scheme governing particle positions and feature values, we derived a kinetic model for the system. In the quasi-invariant regime, we obtained a mean-field description in the form of a Fokker–Planck-type equation, combining a nonlocal aggregation–diffusion operator in space with a nonlocal transport term in the feature variable. By introducing distinct time scales for the different interactions, we further derived a first-order macroscopic model describing the evolution of the feature density. Leveraging this reduced-complexity formulation, we developed a data-driven optimization procedure to identify optimal model parameters such that the asymptotic feature distribution matches the Ground Truth Segmentation Mask. The optimization is performed using consensus-based optimization techniques, which effectively handle the non-convexity of the loss function. Numerical experiments on images with different noise configurations demonstrate the robustness of the proposed approach across various noise types and intensities, as well as its ability to accurately recover the expected segmentation masks. Potential future research directions is to couple the presented technique with learning-based methods as way to obtain the optimal interactions for unsegmented images. 

\section*{Acknowledgements}
The research underlying this paper has been undertaken within the activities of the GNFM group
of INdAM (National Institute of High Mathematics). All the authors acknowledge partial support from the PRIN2022PNRR project No.P2022Z7ZAJ, European Union - NextGenerationEU. M.Z. acknowledges partial support by ICSC - Centro Nazionale di Ricerca in High Performance Computing, Big Data and Quantum Computing, funded by European Union - NextGenerationEU.

\end{document}